%% file: mulut.tex
\documentclass[10pt,journal,compsoc]{IEEEtran}

\usepackage[nocompress]{cite}
\usepackage{multirow}
\usepackage{booktabs}
\usepackage{threeparttable}

\usepackage{array}
\newcolumntype{H}{>{\setbox0=\hbox\bgroup}c<{\egroup}@{}}
\newcommand{\PreserveBackslash}[1]{\let\temp=\\#1\let\\=\temp}
\newcolumntype{C}[1]{>{\PreserveBackslash\centering}p{#1}}
\newcolumntype{R}[1]{>{\PreserveBackslash\raggedleft}p{#1}}
\newcolumntype{L}[1]{>{\PreserveBackslash\raggedright}p{#1}}
\PassOptionsToPackage{table,xcdraw}{xcolor}
\usepackage[]{xcolor}

\usepackage{comment} 

\usepackage{graphicx}
\usepackage{blindtext}

\usepackage{adjustbox}
\usepackage{wrapfig}

\usepackage[colorlinks,linkcolor=black,anchorcolor=black,citecolor=black]{hyperref}
\usepackage{subfigure}

\usepackage{amsmath}
\usepackage{amssymb}

\begin{document}

\title{Toward DNN of LUTs: Learning Efficient \\ Image Restoration with Multiple Look-Up Tables}

\author{Jiacheng~Li, 
        Chang~Chen\thanks{J.~Li, Z.~Cheng, and Z.~Xiong are with the University of Science and Technology of China, Hefei, Anhui 230052, China. E-mail: \{jclee, mywander\}@mail.ustc.edu.cn, zwxiong@ustc.edu.cn.},
        Zhen~Cheng\thanks{C.~Chen is with Noah's Ark Lab, Huawei Technologies. E-mail: changc25@huawei.com.},
        and~Zhiwei~Xiong
}

\markboth{Journal of \LaTeX\ Class Files,~Vol.~14, No.~8, August~2015}%
{Li \MakeLowercase{\textit{et al.}}: Toward DNN of LUTs: Learning Efficient Image Restoration with Multiple Look-Up Tables}

\IEEEtitleabstractindextext{%
\begin{abstract}
\input{parts/abstract.tex}

\end{abstract}

}

\maketitle

\IEEEdisplaynontitleabstractindextext

\IEEEpeerreviewmaketitle

\input{parts/introduction.tex}
\input{parts/related.tex}

\input{parts/background.tex}

\input{parts/method.tex}
\input{parts/experiments.tex}

\input{parts/conclusion.tex}



\ifCLASSOPTIONcaptionsoff
  \newpage
\fi

\bibliographystyle{IEEEtran}
\bibliography{IEEEabrv,bib}

\end{document}

%% file: parts/abstract.tex
The widespread usage of high-definition screens on edge devices stimulates a strong demand for efficient image restoration algorithms.
The way of caching deep learning models in a look-up table (LUT) is recently introduced to respond to this demand.
However, the size of a single LUT grows exponentially with the increase of its indexing capacity, which restricts its receptive field and thus the performance.
To overcome this intrinsic limitation of the single-LUT solution, we propose a universal method to construct multiple LUTs like a neural network, termed MuLUT.
Firstly, we devise novel complementary indexing patterns, as well as a general implementation for arbitrary patterns, to construct multiple LUTs in parallel.
Secondly, we propose a re-indexing mechanism to enable hierarchical indexing between cascaded LUTs.
Finally, we introduce channel indexing to allow cross-channel interaction, enabling LUTs to process color channels jointly.
In these principled ways, the total size of MuLUT is linear to its indexing capacity, yielding a practical solution to obtain superior performance with the enlarged receptive field.
We examine the advantage of MuLUT on various image restoration tasks, including super-resolution, demosaicing, denoising, and deblocking. 
MuLUT achieves a significant improvement over the single-LUT solution, \emph{e.g.}, up to 1.1dB PSNR for super-resolution and up to 2.8dB PSNR for grayscale denoising, while preserving its efficiency, which is 100$\times$ less in energy cost compared with lightweight deep neural networks. 
Our code and trained models are publicly available at \url{https://github.com/ddlee-cn/MuLUT}.

%% file: parts/introduction.tex
\IEEEraisesectionheading{\section{Introduction}\label{sec:intro}}

Image restoration aims to generate high-quality (HQ) visual data with high-frequency details from low-quality (LQ) observations (\emph{e.g.}, downscaled, noisy, and compressed images). Image restoration algorithms enjoy wide applications, ranging from visual quality enhancement \cite{DBLP:conf/cvpr/IraniP92,DBLP:conf/icmcs/XiongSW08}, digital holography \cite{DBLP:journals/ejasp/Zhang06}, satellite imaging \cite{DBLP:journals/tgrs/TatemLAN01}, medical imaging \cite{DBLP:journals/cj/Greenspan09}, and gaming \cite{amdsr,nvidiasr}. Moreover, besides improving image quality, image restoration helps in many other computer vision tasks, \emph{e.g.}, human face recognition \cite{DBLP:conf/cvpr/WangLZS21}, scene understanding \cite{DBLP:conf/fgr/ZhangLX18}, and autonomous driving \cite{DBLP:journals/ijcv/XueCWWF19}.

Recent methods based on deep neural networks (DNNs) \cite{DBLP:conf/eccv/DongLHT14,DBLP:conf/cvpr/KimLL16a,DBLP:conf/eccv/DongLT16,DBLP:conf/cvpr/LimSKNL17,DBLP:journals/tip/ZhangZCM017,DBLP:journals/tip/ZhangZZ18,DBLP:conf/eccv/WangYWGLDQL18,DBLP:journals/pami/ChenXTZW20} have made impressive progress in restoration performance, thanks to their scalability and flexibility from constructing elementary building blocks like convolutional layers. However, superior performance is usually obtained at a cost of heavy computational burden. Although this can be alleviated by elaborate network structures or dedicated computing engines (\emph{e.g.}, GPU and NPU), the hardware cost and power consumption still limit the deployment of existing deep restoration networks. Specifically, the growing number of high-definition screens on edge devices (\emph{e.g.}, smartphones and televisions) calls for a practical restoration solution.

On the other hand, in the image processing pipeline \cite{DBLP:journals/tip/MukherjeeM08,DBLP:journals/tog/MantiukDK08,DBLP:journals/pami/KimLLSLB12}, look-up table (LUT) is a widely-used mapping operator, especially for color manipulation. For sRGB color-wise mapping, the source colors and the corresponding target colors are stored in index-value pairs in a LUT. This way, each pixel can be directly mapped to the target color with highly efficient memory access. An emerging research, SR-LUT \cite{DBLP:conf/cvpr/JoK21}, adopts LUT to image super-resolution by building spatial-wise mapping between low-resolution (LR) patches and high-resolution (HR) patches. Specifically, SR-LUT utilizes a single LUT to cache the exhaustive HR patch values for later retrieval, which are computed in advance by a learned super-resolution network. At inference time, the LR patches sampled from neighboring pixels are compared with indexes in the LUT, and the cached HR patch values are retrieved. This contributes significantly to the power efficiency and inference speed, making SR-LUT a distinct solution other than existing lightweight super-resolution networks \cite{DBLP:conf/eccv/DongLT16,DBLP:conf/eccv/AhnKS18,DBLP:conf/eccv/LiYLZZYJ20}. However, in practice, the size of LUT is limited by the on-device memory. For a single LUT, the size grows \textbf{exponentially} as the dimension of indexing entries (\emph{i.e.}, indexing capacity) increases. This imposes a restriction on the indexing capacity as well as the corresponding receptive field (RF) of the super-resolution network to be cached, which is the main obstacle to performance improvement.


In this paper, we embrace the merits of LUT and propose a universal method to overcome its intrinsic limitation, by enabling the cooperation of \textbf{Mu}ltiple \textbf{LUT}s, termed MuLUT. Inspired by the construction of a common DNN, we propose three fundamental ways to construct LUTs in the spatial, depth, and channel dimensions. 1) In the spatial dimension, we devise novel \textit{complementary indexing} patterns as well as a general implementation for realizing arbitrary patterns, to construct LUTs in a parallel manner. 2) In the depth dimension, we enable \textit{hierarchical indexing} between cascaded LUTs, by proposing a re-indexing mechanism to link between LUTs from different hierarchies. 3) In the channel dimension, we introduce \textit{channel indexing}, where channel-wise LUTs are inserted between spatial-wise LUTs, to allow cross-channel interaction in processing color images. Besides, we propose a LUT-aware finetuning strategy to mitigate the performance drop after converting the learned neural network to LUTs. In these principled ways, MuLUT takes a single LUT as a network layer and works as a network of LUTs. Thus, the RF of MuLUT can be enlarged on demand. Meanwhile, the total size of MuLUT is \textbf{linear} to its indexing capacity, yielding a practical solution to obtain superior performance at high efficiency.

Extensive experiments demonstrate a clear advantage of our proposed MuLUT compared with the single-LUT solution. On five super-resolution benchmarks, MuLUT achieves up to 1.1dB PSNR improvement over SR-LUT, approaching the performance of the lightweight FSRCNN model \cite{DBLP:conf/eccv/DongLT16}. At the same time, MuLUT preserves the efficiency of the single-LUT solution. For example, the energy cost is about 100 times less than that of FSRCNN. To demonstrate the versatility of MuLUT, we also extend MuLUT to more image restoration tasks, including demosaicing, denoising, and deblocking. Although a single LUT can be built for these tasks similar to SR-LUT, it yields inferior performance due to the restricted RF. Instead, MuLUT enlarges the RF on demand with flexible architecture choices. As a result, compared with the single-LUT solution, MuLUT achieves over 6.0dB cPSNR gain for demosaicing, up to 2.8dB PSNR gain for denoising, and up to 0.85dB PSNR-B gain for deblocking.

A preliminary version of MuLUT appears in \cite{mulut_eccv}, where it is dedicated to efficient super-resolution and also applied to demosaicing. In this paper, we extend MuLUT substantially towards a universal method for efficient image restoration in the following aspects. 1) In the spatial dimension, we devise more indexing patterns to further integrate neighboring information and enlarge the RF. Moreover, we formulate a general implementation to support arbitrary indexing patterns, which increases the flexibility of MuLUT. 2) In the newly introduced channel dimension, we propose to insert channel-wise LUTs between spatial-wise LUTs to allow cross-channel interaction. In this way, MuLUT can cache more complicated neural networks for color image processing. 3) We extend MuLUT to more image restoration tasks, including denoising and deblocking, showing significant advantages over the single-LUT solution and the versatility of our method. 4) We provide a more thorough literature review, a clearer motivation of MuLUT, more comprehensive experimental settings and results, as well as in-depth discussions on open questions.

With the above extensions, the contributions of this work can be summarized as follows:

1) We devise a universal method, named MuLUT, for efficient image restoration by enabling the cooperation of multiple LUTs. Our method treats a single LUT as a network layer and constructs multiple LUTs like a common DNN.

2) We overcome the intrinsic limitation of the single-LUT solution by introducing complementary indexing, hierarchical indexing, and channel indexing. Our method enlarges the receptive field at a linearly growing cost, instead of exponentially. 

3) We propose a series of techniques, \emph{i.e.}, the general implementation for arbitrary patterns, the re-indexing mechanism, and the LUT-aware finetuning strategy, to enable multiple LUTs to be effectively learned from data.

4) Extensive experiments on four representative image restoration tasks demonstrate that MuLUT achieves a significant improvement in performance over the single-LUT solution while preserving a clear advantage in efficiency over DNNs, showing its versatility for wide applications on edge devices.



%% file: parts/related.tex
\section{Related Works}

\subsection{Image Restoration}
\noindent\textbf{Super-Resolution.}
Image super-resolution aims to generate a high-resolution image by restoring or enhancing high-frequency details from its low-resolution measurements. Interpolation operators, including nearest, bilinear, and bicubic \cite{Keys1981CubicCI}, are highly efficient and widely used on edge devices, but they often produce blurry results because the interpolation weights are calculated without considering the local structure inside the image. Example-based methods leverage a dataset of LR-HR image patch pairs \cite{DBLP:journals/ijcv/FreemanPC00,DBLP:journals/tmm/XiongXSW13,DBLP:conf/iccv/TimofteDG13,DBLP:conf/accv/TimofteSG14}, or exploit self-similarity inside the LR image \cite{DBLP:conf/iccv/GlasnerBI09}, among which sparse coding methods learn a compact representation of the patches, showing promising results \cite{DBLP:journals/tip/YangWHM10,DBLP:conf/cas/ZeydeEP10}. Other super-resolution methods based on random forests \cite{DBLP:conf/cvpr/SchulterLB15}, gradient filed sharpening \cite{DBLP:journals/tip/SongXLXWG18}, and displacement field \cite{DBLP:journals/tip/WangWP14} are also explored. Nevertheless, these classical super-resolution methods suffer from either unsatisfying visual quality or time-consuming optimization.
With the rise of DNN methods, the community has made impressive progress in the task of super-resolution \cite{DBLP:conf/eccv/DongLHT14,DBLP:conf/cvpr/KimLL16a,DBLP:conf/eccv/DongLT16,DBLP:conf/cvpr/LimSKNL17,DBLP:conf/eccv/AhnKS18,DBLP:conf/cvpr/ZhangTKZ018,DBLP:conf/eccv/ZhangLLWZF18,DBLP:conf/eccv/WangYWGLDQL18,DBLP:conf/cvpr/ChenXTZW19,DBLP:conf/cvpr/XiaoFHCX21,DBLP:conf/cvpr/ChengXC0Z21}. 



\noindent\textbf{Demosaicing.} 
Image demosaicing aims to produce colored observation from linear responses of light sensors inside the camera. It can be viewed as a super-resolution problem with a particular color pattern, typically the Bayer pattern. Interpolation-based methods like nearest and bilinear can also be used in image demosaicing. However, they tend to produce artifacts in the region with high-frequency signal changes. Classical methods taking advantage of self-similarity inside the image \cite{DBLP:journals/tip/BuadesCMS09,DBLP:journals/tip/DuranB14} or relying on an optimization process \cite{DBLP:journals/tip/ZhangW05,DBLP:journals/tip/Jeon013} are proposed, but they require an excessive amount of computing time. Recently, DNN methods have been introduced to take advantage of powerful representations learned from large-scale datasets \cite{DBLP:journals/tog/Durand16a,DBLP:journals/tip/KokkinosL19}. 


\input{pics/fig1_srlut_recap.tex}

\noindent\textbf{Denoising.} 
Image denoising is a long-lived task in low-level vision. Early methods treat the denoising task as an image filtering problem, producing results by modifying transform coefficients \cite{DBLP:conf/icip/SimoncelliA96} or averaging neighboring image pixels \cite{DBLP:journals/pami/PeronaM90}. Tremendous methods are explored to model image priors, including self-similarity \cite{DBLP:journals/ijcv/BuadesCM08,DBLP:journals/pami/ChenP17}, sparse representation \cite{DBLP:conf/iccv/MairalBPSZ09}, gradient filed \cite{DBLP:conf/cvpr/WeissF07}, and Markov random field \cite{DBLP:journals/ijcv/RothB09}. Among them, methods based on self-similarity such as BM3D \cite{DBLP:journals/tip/DabovFKE07} and WNNM \cite{DBLP:conf/cvpr/GuZZF14} show promising results, but the searching process for similar patches is time-consuming. Recently, DNN methods have made significant progress in image denoising, showing the advantage of learning from data \cite{DBLP:journals/tip/ZhangZCM017,DBLP:journals/tip/ZhangZZ18,DBLP:journals/pami/ChenXTZW20,DBLP:conf/cvpr/ChenXL020,DBLP:journals/pami/ZhangTKZF21,DBLP:conf/cvpr/ZamirA0HK0021,DBLP:conf/iccvw/LiangCSZGT21}. 


\noindent\textbf{Deblocking.} 
Image deblocking, also referred to as compression artifacts reduction, aims to reduce the blocking artifacts caused by the inexact approximations in lossy compression (\emph{e.g.}, JPEG, WebP, and HEIF). Early methods based on filtering \cite{DBLP:journals/tcsv/ListJLBK03,DBLP:journals/spic/WangZL13}, frequency-domain transformation \cite{DBLP:journals/tip/FoiKE07}, and optimization \cite{DBLP:journals/tip/SunC07} are explored. Recently, DNN methods are introduced to the task of image deblocking, making great progress \cite{DBLP:conf/iccv/DongDLT15,DBLP:conf/eccv/GuoC16,DBLP:conf/eccv/Ehrlich0LS20}. 



\subsection{Efficient image restoration} 
DNN methods obtain impressive performance in image restoration tasks, but generally require dedicated computing engines (\emph{e.g.}, GPU and NPU) with high power consumption. Thus, many efforts for efficient image restoration have been conducted. Taking super-resolution as an example, researchers elaborately design lightweight networks, including ESPCN \cite{DBLP:conf/cvpr/ShiCHTABRW16}, FSRCNN \cite{DBLP:conf/eccv/DongLT16}, CARN-M \cite{DBLP:conf/eccv/AhnKS18}, and IMDN \cite{DBLP:conf/cvpr/HuiWG18}, to name a few. General network compression methods like quantization \cite{DBLP:conf/eccv/LiYLZZYJ20,DBLP:journals/tip/ZhangLWZ21}, neural architecture search \cite{DBLP:conf/eccv/LeeDAVK0L20}, network pruning \cite{DBLP:conf/eccv/LiGZGT20}, and AdderNet \cite{DBLP:conf/cvpr/ChenWXSX0X20,DBLP:conf/cvpr/Song0C0XT21} have also been explored for efficient super-resolution. Beyond super-resolution, DNN methods for efficient denoising \cite{DBLP:journals/tip/ZhangZZ18,DBLP:conf/iccv/GuLGT19} have also been reported. However, a substantial computational burden is still required for DNNs to deal with numerous floating-point operations, which limits their application on edge devices with limited hardware resources and energy supply.

Most recently, Jo \emph{et al.} propose SR-LUT \cite{DBLP:conf/cvpr/JoK21} as a new efficient solution. They train a deep super-resolution network with a restricted RF and then cache the output values of the learned super-resolution network to a LUT. At test time, they retrieve the pre-computed HR output values from the LUT for the query LR input pixels. Different from the color-to-color 3D LUTs that are widely used in the image processing pipeline \cite{DBLP:journals/tip/MukherjeeM08,DBLP:journals/tog/MantiukDK08,DBLP:journals/pami/KimLLSLB12} and the image enhancement task \cite{DBLP:journals/pami/ZengCLCZ22,DBLP:conf/iccv/Wang0PMWSY21}, SR-LUT builds spatial-wise mapping between a local patch and its high-frequency counterpart, resulting in a patch-to-patch 4D LUT. However, a single LUT yields inferior performance due to the restriction of the dimension of indexing entries, which equals to the RF of the learned super-resolution network. This is proved to be critical for super-resolution \cite{DBLP:conf/cvpr/GuD21}. Our proposed MuLUT in \cite{mulut_eccv} overcomes the intrinsic limitation of SR-LUT by enabling the cooperation of multiple LUTs. A concurrent work, SPLUT \cite{Ma2022LearningSL}, also improves SR-LUT by cascading more LUTs. However, since the indexing patterns of SPLUT are similar to SR-LUT, a large number of LUTs is required to obtain a modest RF. With novel pattern designs in complementary indexing, MuLUT is able to enlarge RF more effectively. In this paper, we further extend MuLUT to more image restoration tasks with the newly introduced channel indexing and more flexible pattern designs in complementary indexing, which serves as a universal method toward ``DNN of LUTs'' for efficient image restoration.



%% file: pics/fig1_srlut_recap.tex
\begin{figure*}[t]
    \definecolor{myblue}{RGB}{180, 199, 231}
    \definecolor{mygreen}{RGB}{169, 209, 142}
    \centering
    \includegraphics[width=\textwidth]{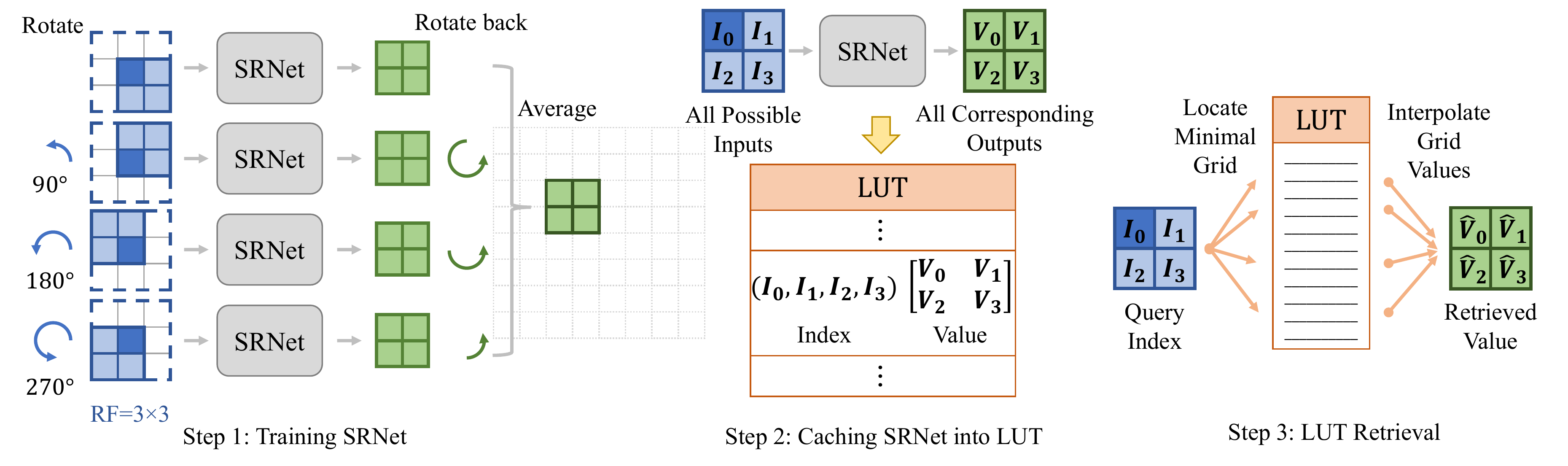}

    \caption{Recap of SR-LUT \cite{DBLP:conf/cvpr/JoK21}. Firstly, a deep super-resolution network is trained. Next, a look-up table (LUT) is obtained by caching the output values of the learned deep super-resolution network by traversing all possible inputs. Finally, the prediction results are retrieved by locating query indexes to the minimum grid and interpolating the pre-computed grid values from the LUT. The indexing entries and corresponding HR values of a 4D LUT for $2\times$ super-resolution are marked in \colorbox{myblue}{blue} and \colorbox{mygreen}{green}, respectively. The actual receptive area with the rotation ensemble trick are depicted with dashed lines.} 
    \label{fig:srlut_recap}
\end{figure*}

%% file: parts/background.tex
\section{Motivation}
\subsection{Preliminary}

\setcounter{figure}{2}

\input{pics/fig3_overview.tex}
\setcounter{figure}{1}    
\input{pics/fig2_motivation.tex}
\setcounter{figure}{3}    

LUT is a widely-used mapping operator, especially for color manipulation modules in the image processing pipeline \cite{DBLP:journals/tip/MukherjeeM08,DBLP:journals/tog/MantiukDK08,DBLP:journals/pami/KimLLSLB12}. A LUT is composed of records of indexes and values, which play as lookup indexing entries and interpolation candidates at the inference time, respectively. These paired indexes and values can be stored in the on-device memory, resulting in high execution efficiency. 

Recently, Jo \textit{et al.} proposed SR-LUT, adopting LUT to the task of single-image super-resolution \cite{DBLP:conf/cvpr/JoK21}. As illustrated in Fig.~\ref{fig:srlut_recap}, there are three steps to obtain a LUT for SR. Firstly, A deep super-resolution network with a restricted RF is trained. Then, the output values of the trained super-resolution network are cached into a LUT by traversing all possible inputs and pre-computing all corresponding outputs. The LUT is uniformly sampled to reduce size. Finally, the HR predictions are obtained by locating the minimum grid of the LR input pixels in the sampled LUT and interpolating cached grid values. This way, the exhaustive results of a deep super-resolution network with a $2 \times 2$ RF size can be cached and retrieved by a 4D LUT. A rotation ensemble strategy is also used to further enlarge the RF. This process can be formulated as
\begin{equation}
    \hat{HR}=\frac{1}{4} \sum_{j=0}^{3} R_{j}^{-1}\left(f\left(R_{j}\left(LR\right)\right)\right),
 \end{equation}
\noindent where $f$ denotes the forward function, $R_j$ and $R_j^{-1}$ denote the $j$ times of $\mathtt{rot90}()$ and its inverse operation, respectively. With the rotation ensemble strategy, the RF size of a 4D LUT can be enlarged from $2 \times 2$ pixels to $3 \times 3$. By caching a trained super-resolution network to a LUT, SR-LUT achieves comparable efficiency with bicubic interpolation.

\input{pics/fig4_pattern.tex}

\subsection{Problem and Our Solution}


Due to a relatively small RF, the performance of SR-LUT is still much inferior to lightweight DNNs like FSRCNN \cite{DBLP:conf/eccv/DongLT16}. Typically, better performance can be expected by enlarging its RF \cite{DBLP:conf/cvpr/GuD21}. On the other hand, in SR-LUT, LUT is adopted to avoid the online computation of a complex function, by caching the pre-computed results. To this end, one needs to traverse \textit{all possible combinations} of input values for the offline pre-computation. Due to the exhaustive combination, the size of cached results in a single LUT grows exponentially with respect to the increasing input dimension. For an 8bit LUT, whose indexes and values are 8bit integers, its size $S$ can be calculated as
\begin{equation} \label{eq:lut_size}
   S = (2^{8-q}+1)^n \times m \mathtt{B},
\end{equation}
\noindent where $q$ is the uniform sampling interval, $n$ the index dimension of the LUT, and $m$ the number of values for each record, \emph{e.g.}, $m=4\times4=16$ for $4 \times$ super-resolution. Thus, the exponential growth of the LUT size makes it unacceptable to obtain better performance by increasing the dimension of a single LUT.

In this work, we address this exponential disaster by enabling the cooperation of multiple LUTs, instead of staying with a single LUT. As shown in Fig.~\ref{fig:motivation}(a), our method reduces the exponential growth of the LUT size to a linear growth as the number of covered pixels increases. Consequently, as illustrated in Fig.~\ref{fig:motivation}(b), at a comparable energy cost, MuLUT outperforms SR-LUT by a large margin, approaching similar restoration performance compared with FSRCNN \cite{DBLP:conf/eccv/DongLT16} but at two orders of magnitude lower energy cost.

%% file: pics/fig3_overview.tex
\begin{figure*}[t]
    \begin{center}
        \subfigure[Cooperation of multiple LUTs with complementary, hierarchical, and channel indexing.]{
        \includegraphics[width=0.9\textwidth]{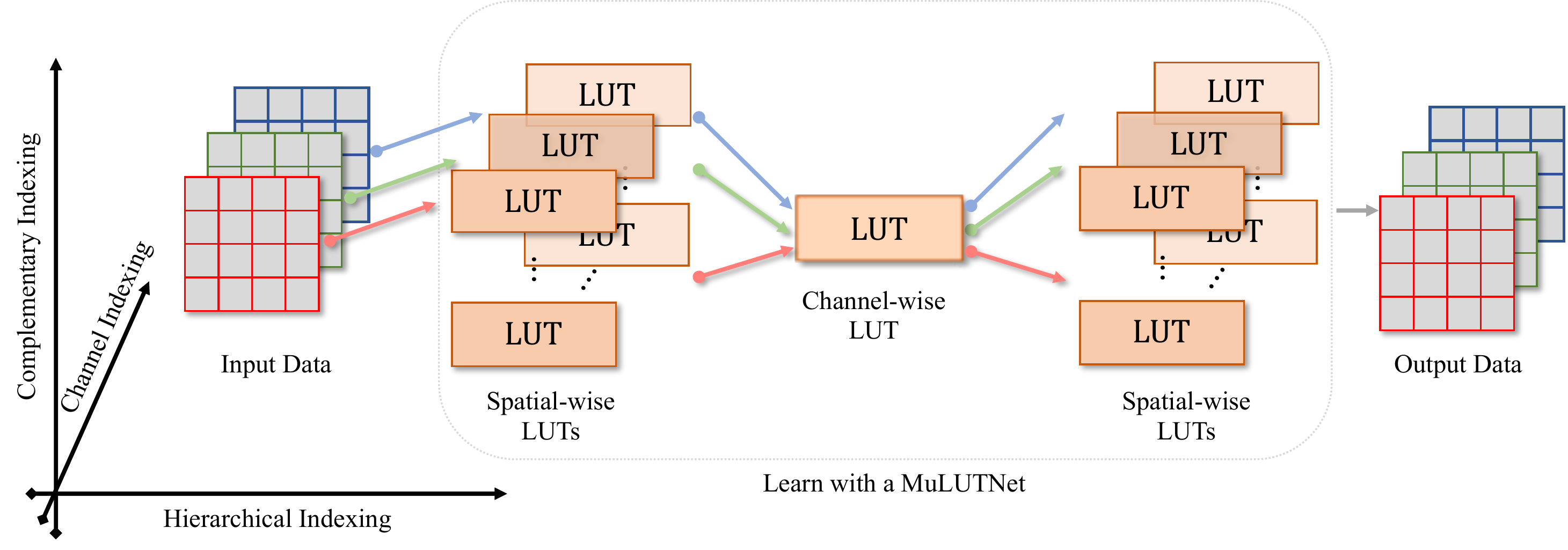}}

        \subfigure[Different types of the MuLUT Block in a MuLUTNet.]{
        \includegraphics[width=0.9\textwidth]{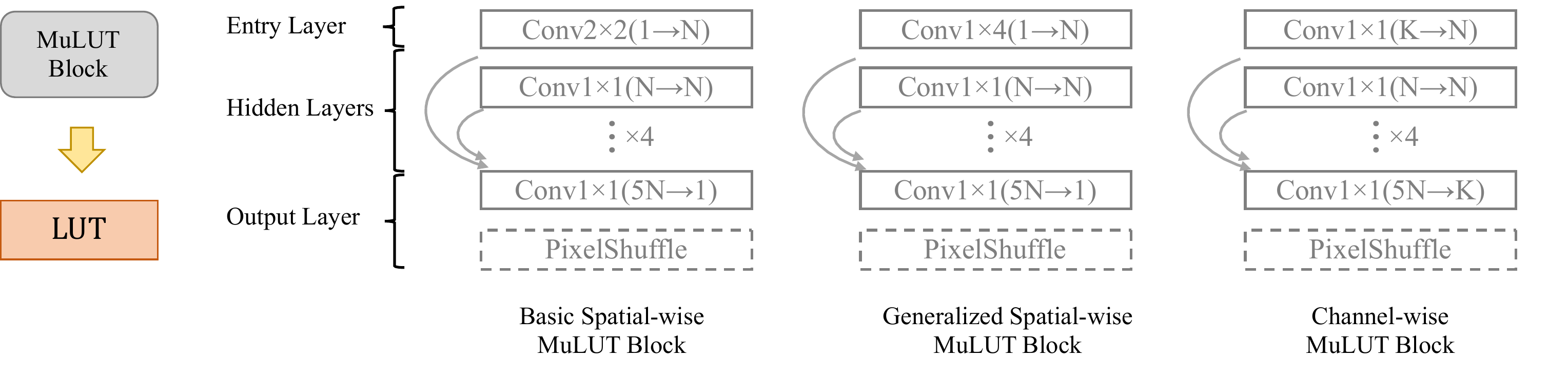}}
    \end{center}
    \caption{Overview of MuLUT. (a) With complementary, hierarchical, and channel indexing, LUTs can be constructed in a general and flexible way like neural networks. These LUTs are learned with a MuLUTNet, which is composed of multiple MuLUT blocks in a shared structure. After training, the inputs and outputs of each MuLUT block are cached into a LUT, while the computation graph is retained. (b) We design different types of MuLUT blocks to construct the learnable MuLUTNet. The convolution layer is denoted in the format of $\mathtt{kernel\_width} \times \mathtt{kernel\_height}(\mathtt{input\_channel} \rightarrow \mathtt{output\_channel})$. The connecting lines denotes the dense connection \cite{DBLP:conf/cvpr/HuangLMW17}.}
    \label{fig:overview}
\end{figure*}

%% file: pics/fig2_motivation.tex
\begin{figure}[t]
    \begin{center}
    \subfigure[]{
        \includegraphics[width=0.45\columnwidth]{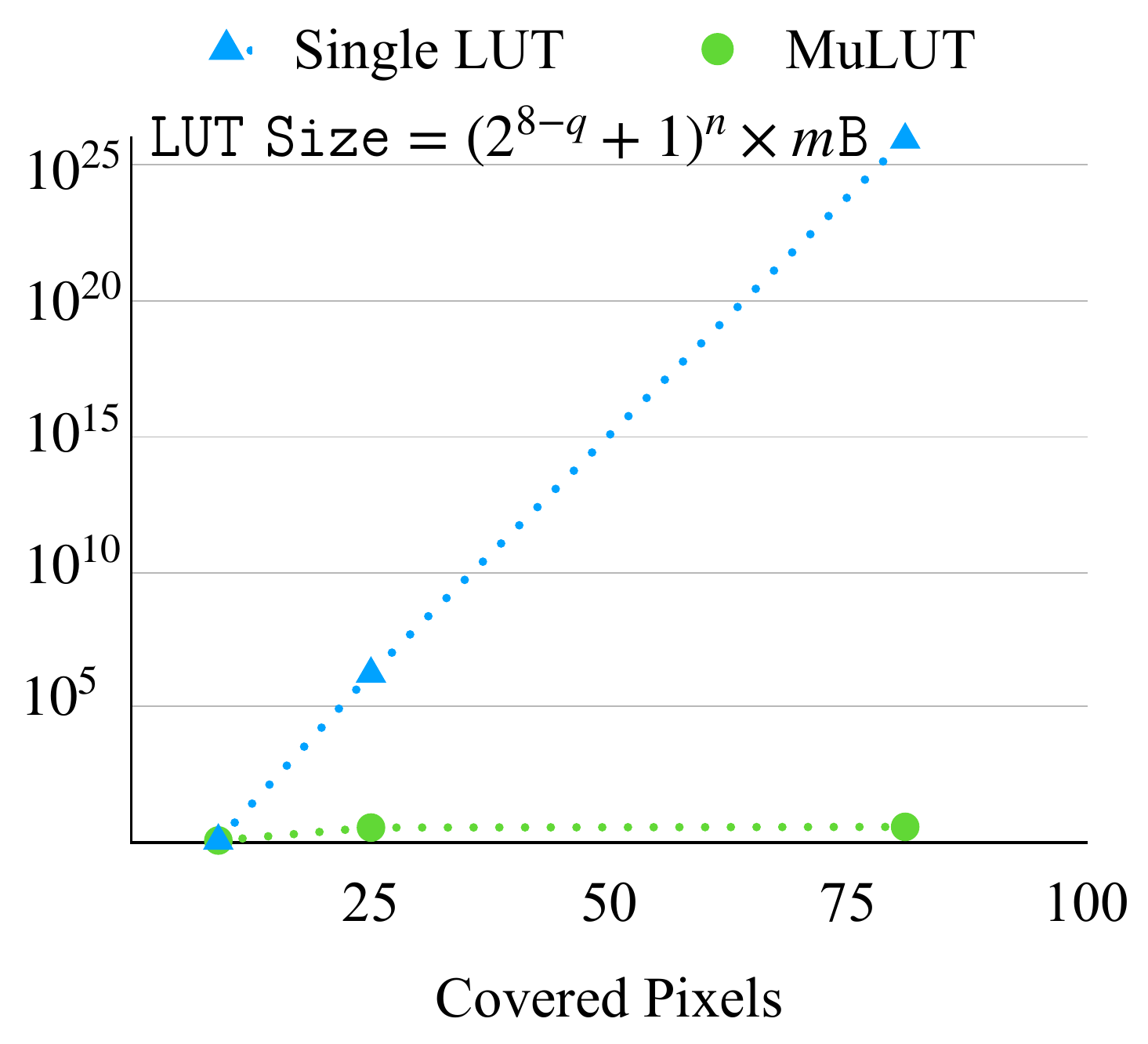}}
    \subfigure[]{
        \includegraphics[width=0.45\columnwidth]{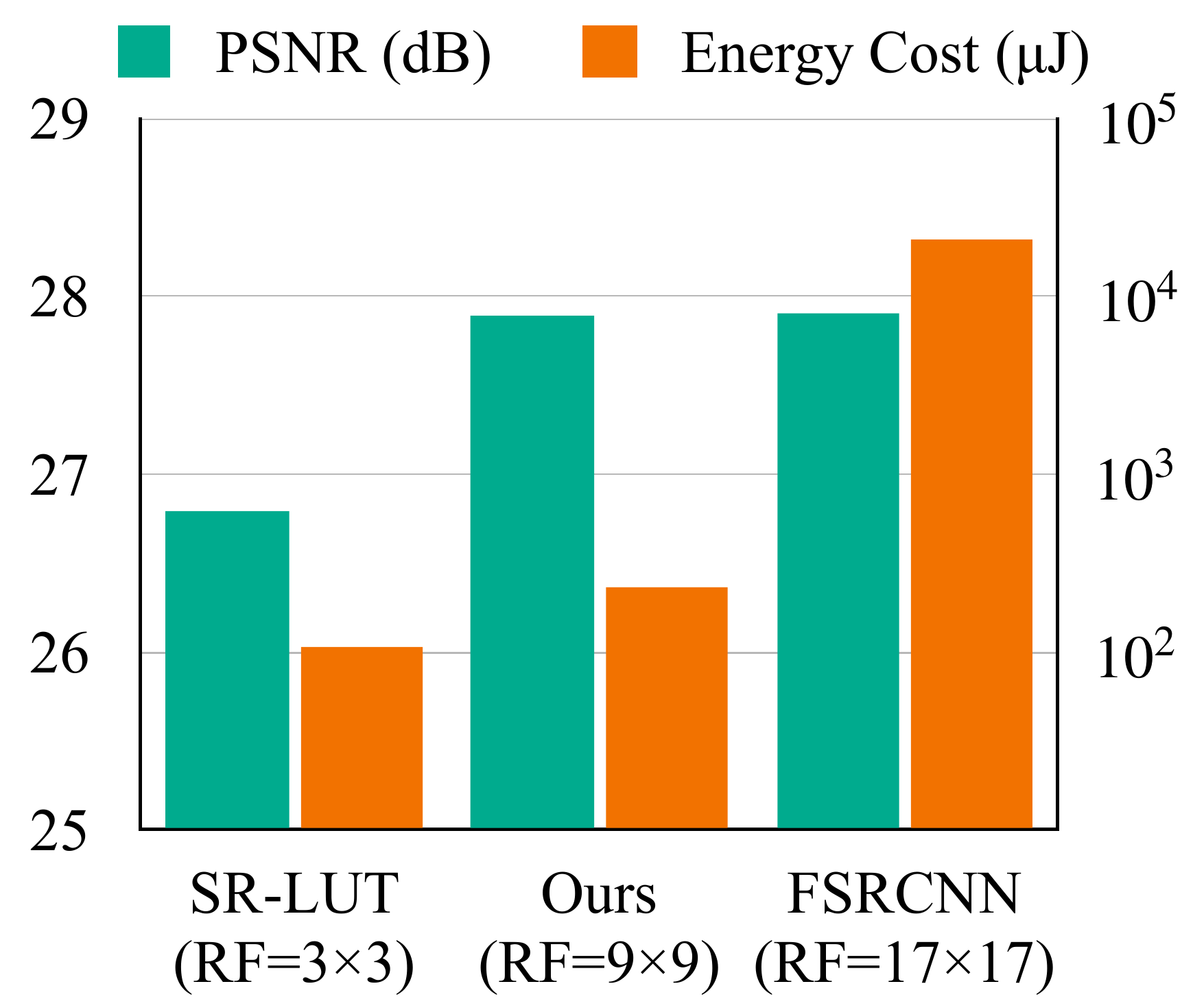}}
    \end{center}
    \caption{(a) For a single LUT, its storage size grows exponentially as the number of covered pixels increases. Our method provides a simple but effective solution to avoid this exponential growth. (b) By enabling cooperation of multiple LUTs, we enlarge the RF from $3 \times 3$ to $9 \times 9$, resulting in a significant performance improvement over SR-LUT while preserving its efficiency. The PSNR values are evaluated on Manga109 for $4\times$ super-resolution.}
    \label{fig:motivation}
\end{figure}

%% file: pics/fig4_pattern.tex
\begin{figure*}[t]
    \centering
    \includegraphics[width=\textwidth]{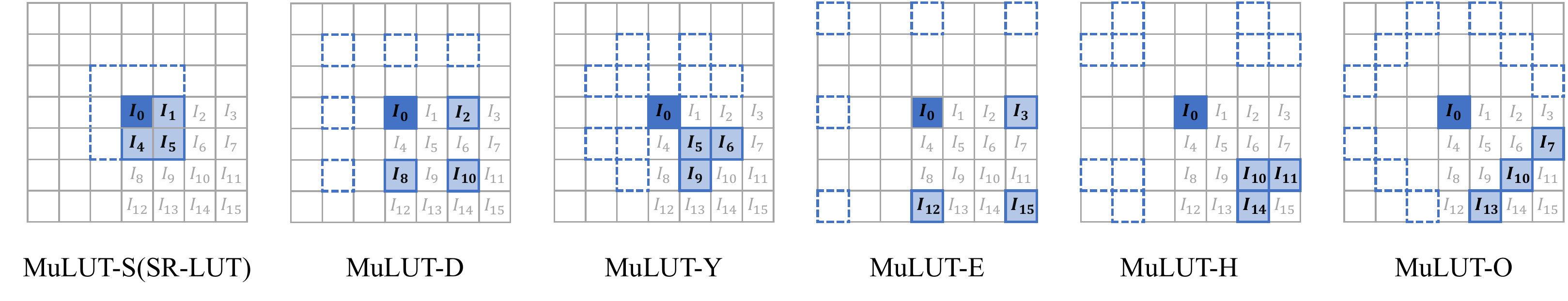}
    \caption{Complementary indexing of multiple LUTs. With the proposed novel indexing patterns, MuLUT involves more pixels than a single SR-LUT. For example, With MuLUT-S, MuLUT-D, and MuLUT-Y, the $5 \times 5$ area around $I_0$ is fully covered. The covered pixels with the rotation ensemble trick are marked with dashed boxes.} 
    \label{fig:comp_index}
\end{figure*}

%% file: parts/method.tex
\section{Cooperation of Multiple Look-Up Tables}


\mathchardef\mhyphen="2D 
\newcommand\luts{\mathop{LUT_S}}
\newcommand\lutd{\mathop{LUT_D}}
\newcommand\luty{\mathop{LUT_Y}}
\newcommand\lutx{\mathop{LUT}} 

\newcommand\lutone{\mathop{LUT^{(1)}}}
\newcommand\luttwo{\mathop{LUT^{(2)}}}
\newcommand\lutN{\mathop{LUT^{(N)}}}
\newcommand\lutM{\mathop{LUT^{(M)}}}





\subsection{Overview}


Inspired by the construction of a common DNN, we treat a single LUT as an elementary component and construct multiple LUTs in the spatial, depth, and channel dimensions. Specifically, as illustrated in Fig.~\ref{fig:overview}(a), we propose three fundamental ways, \emph{i.e.}, complementary indexing, hierarchical indexing, and channel indexing, to generalize a single LUT to MuLUT, whose RF can be effectively enlarged by constructing multiple elementary components as a neural network. For clarity, we term the elementary component as a ``MuLUT Block'' and the resulting network as a ``MuLUTNet''. As shown in Fig.~\ref{fig:overview}(b), we design different types of MuLUT blocks, including two spatial-wise types and one channel-wise type. By parallelizing and cascading these MuLUT blocks, the RF of the MuLUTNet increases, while the total size of the cached LUTs grows linearly instead of exponentially. In this way, MuLUT equips with a much larger indexing capacity without introducing the enormous cost of storage and computation.


In the following sections, we will detail the proposed indexing techniques to enable the cooperation of multiple LUTs, as well as a LUT-aware finetuning strategy.


\subsection{Constructing LUTs Spatially}

\input{pics/fig5_gmblock.tex}


\input{pics/fig6_hindex.tex}
The first cooperation in MuLUT is parallelizing LUTs with complementary indexing. For image restoration tasks, the surrounding pixels provide critical information to restore the high-frequency details, making it essential to cover as many as input pixels for restoration models. Thus, we construct multiple LUTs with different indexing patterns in parallel, which are carefully designed to complement each other. For 4D LUTs, besides the standard indexing pattern introduced in SR-LUT (named MuLUT-S here), we devise more novel indexing patterns (MuLUT-D, MuLUT-Y, MuLUT-E, MuLUT-H, and MuLUT-O), as shown in Fig.~\ref{fig:comp_index}. These patterns cover complementary pixels, for example, the indexing pixels of MuLUT-S, MuLUT-D, and MuLUT-Y are $(I_{0}, I_{1}, I_{3}, I_{4})$, $(I_{0}, I_{2}, I_{6}, I_{8})$, and $(I_{0}, I_{4}, I_{5}, I_{7})$, respectively. Our design covers the whole $5 \times 5$ area with these three types of MuLUT blocks working together. Even more pixels can be covered by further involving MuLUT-E, MuLUT-H, and MuLUT-O blocks. Correspondingly, the MuLUTNet is designed to be with multiple branches, where the parallel MuLUT blocks with complementary pixels are jointly trained. The cached LUTs are then retrieved in parallel, after which their predictions are averaged. Generally, for anchor $I_0$, the corresponding output values $V$ are obtained by
\begin{equation}
    V = \frac{\sum^{N}_{1}{\lutx[I_*]}}{N},
\end{equation}
where $N$ denotes the number of parallelized LUTs, and $LUT[\cdot]$ denotes the lookup and interpolation process in LUT retrieval, as illustrated in step 3 of Fig.~\ref{fig:srlut_recap}. 

In practice, the MuLUT-S, MuLUT-D, and MuLUT-E blocks can be implemented with standard convolutions, where the MuLUT-D block equips with an entry convolution layer with a dilation size of 2 and the MuLUT-E block with 3. But the MuLUT-Y, MuLUT-H, and MuLUT-O blocks cannot be readily implemented with standard convolutions. Thus, we propose a general implementation to support arbitrary indexing patterns, as illustrated in Fig.~\ref{fig:gmblock}. Precisely, we first unfold the input image by extracting patches with a sliding window. Then, we sample the pixels according to the specified coordinates and reshape these pixels into $1 \times 4$ vectors, which are fed into a standard convolution with a $1 \times 4$ kernel. Thus, as illustrated in Fig.~\ref{fig:overview}(b), our generalized MuLUT block supports arbitrary indexing patterns of a 4D LUT. This implementation is shared between training the generalized MuLUT block and retrieving the cached LUT. With this formulation, one can devise more spatial patterns effortlessly. In summary, with complementary indexing of parallel LUTs, more surrounding pixels are involved to better capture the local structures, which helps to restore the corresponding high-frequency details.

\subsection{Cascading LUTs in Depth}

The second cooperation in MuLUT is cascading LUTs with hierarchical indexing. As illustrated in Fig.~\ref{fig:hierarch_index}~(left), with cascaded LUTs, we conduct the lookup process in a hierarchical manner. The values ($I^{(2)}_*$) in the previous LUT serve as the indexes of the following LUT. This hierarchical indexing process can be formulated as
\begin{equation}
    V = \lutM[\cdot\cdot\cdot\luttwo[\lutone[I_*]]],
\end{equation}
where $M$ denotes the number of cascaded LUTs. From the perspective of the RF, this process is similar to cascading multiple network layers in a DNN. As shown in Fig.~\ref{fig:hierarch_index} (right), cascading two stages of MuLUT blocks enlarges the RF size from $3 \times 3$ to $5 \times 5$. However, the indexes for image data are sampled and stored in the \emph{int8} data type due to LUT size constraint, while training neural networks requires gradients in the \emph{float} data type. Thus, we design a LUT re-indexing mechanism to integrate the behavior of hierarchical indexing in the learning process of the MuLUTNet. Specifically, as shown in Fig.~\ref{fig:hierarch_index}~(right), the prediction values of the previous MuLUT block are quantized to integers in the forward pass while their gradients are retained as floating-point values in the backward pass. This way, the cascaded LUTs can reproduce the performance of the cascaded MuLUT blocks. In practice, as shown in Fig.~\ref{fig:overview}(b), we adopt the dense connections \cite{DBLP:conf/cvpr/HuangLMW17} between hidden layers in the MuLUT blocks to help the convergence of the MuLUTNet with multiple cascaded stages.

\input{pics/fig8_dm.tex}

\input{pics/fig7_mulut_sr.tex}

\subsection{Channel Interaction between LUTs}

The third cooperation in MuLUT is channel indexing. In SR-LUT, the three channels of color images are processed independently, lacking the ability to model cross-color correlations. On the other hand, in the image processing pipeline \cite{DBLP:journals/tip/MukherjeeM08,DBLP:journals/tog/MantiukDK08,DBLP:journals/pami/KimLLSLB12} and image enhancement methods \cite{DBLP:journals/pami/ZengCLCZ22,DBLP:conf/iccv/Wang0PMWSY21}, LUT is adopted to serve as a mapping operator to perform color manipulation. Here, as illustrated in Fig.~\ref{fig:overview}(a), we propose channel indexing, where channel-wise LUTs are integrated with spatial-wise LUTs to allow color channel interaction. The channel-wise LUT is learned with a channel-wise MuLUT block, as shown in Fig.~\ref{fig:overview}(b). Different from spatial-wise LUTs, channel-wise LUTs index pixels across different color channels or timestamps. A $K$ dimensional channel-wise LUT can be converted from a channel-wise MuLUT block with an entry layer of $1 \times 1$ convolution with $K$ input channels. For color image processing, we set $K$ to 3. By inserting channel-wise LUTs between spatial-wise LUTs, we empower multiple LUTs to be constructed like pseudo-3D networks \cite{DBLP:conf/iccv/QiuYM17}.

\subsection{The LUT-aware Finetuning Strategy}
Finally, we propose a finetuning strategy to improve the conversion from the learned neural network to LUTs. In SR-LUT \cite{DBLP:conf/cvpr/JoK21}, due to the constraint of storage, the indexes of a LUT are uniformly sampled to reduce the LUT size, and the nonsampled indexes are approximated with nearest neighbors. Also, an interpolation process is performed to compute final predictions from weighted LUT values during LUT retrieval. The information loss caused by nearest neighbor approximation and interpolation leads to a performance gap between the learned network and the cached LUT (see Table~\ref{tab:ft_abl}). Thus, we propose a LUT-aware finetuning strategy to address this issue. Specifically, we treat the values stored inside LUTs as trainable parameters and finetune them in a similar process to LUT re-indexing, where their forward values are quantized, and their gradients are retained as floating-point in the backward pass. After finetuning, the values inside LUTs are adapted to the sampling and interpolation process. This strategy is universal and serves as a practical improvement to bridge the performance gap between the learned network and the cached LUT for both SR-LUT and MuLUT.

\section{Applications in Image Restoration}

Following the design principles of DNN, MuLUT enables flexible and on-demand construction of multiple LUTs. In the following, we introduce different configurations of MuLUT to be applied in different image restoration tasks. The configurations are also detailed in Table~\ref{tab:config}.



\subsection{Super-Resolution} 

We apply MuLUT to the task of single-image super-resolution to show its advantage in enlarging RF size. Taking the configuration illustrated in Fig.~\ref{fig:mulut_sr} as an intuitive example, we construct a MuLUTNet with 2 cascaded stages, where each stage contains 3 parallel blocks with indexing patterns of ``S'', ``D'', and ``Y'', denoted as MuLUT-SDY-X2. In the second stage, we add the $\mathtt{pixelshuffle}$ operation \cite{DBLP:conf/cvpr/ShiCHTABRW16} to the output layers of MuLUT blocks to enlarge the spatial resolution, \emph{i.e.}, each record of the second stage LUT will contain 4 values for $2 \times$ super-resolution. As shown in Fig.~\ref{fig:mulut_sr}, for an anchor pixel $(x,y)$, it is replaced by 4 pixels in the HR image. MuLUT enlarges the RF from $3 \times 3$ to $9 \times 9$ (9 times larger), while the total size of these LUTs is less than 4 times a single 4D LUT (See Table~\ref{tab:runtime_sr}). In contrast, according to Eq.~\ref{eq:lut_size}, the full size of a 25D LUT with an equivalent $9 \times 9$ RF size is $(2^8)^{25-4}=2^{168}$ times a 4D LUT.


\input{tables/config_table.tex}
\input{tables/energy_table.tex}

\input{tables/comp_table.tex}

\input{pics/fig9_sr_tradeoff.tex}

\subsection{Demosaicing} 

We further show the generalization ability of MuLUT on the task of demosaicing Bayer-patterned images. Image demosaicing can be viewed as a super-resolution problem with a particular color pattern. However, there are grave obstacles to adopting SR-LUT for this task. Two single-LUT baseline solutions are shown in Fig.~\ref{fig:dm}(a) and Fig.~\ref{fig:dm}(b), respectively. In Baseline-A, pixels in a $2 \times 2$ Bayer pattern block are treated as four independent channels, which are processed separately, and then the two green channels are averaged. However, this solution suffers from a subpixel shift of center points due to the misalignment between the HR pixels and the Bayer-patterned sampled ones. In Baseline-B, at a stride of $2$, the $2 \times 2 \times 1$ Bayer-patterned blocks are upsampled into $2 \times 2 \times 3$ colored patches directly. But the limited RF of a single LUT fails to capture the inter-block correlation. In contrast, MuLUT is easy to be adapted to the characteristics of Bayer-patterned images. As shown in Fig.~\ref{fig:dm}(c), MuLUT resembles three color channels like Baseline-B in the first stage, and then integrates the surrounding pixels with three indexing patterns in the second stage. Finally, a channel-wise MuLUT block is cascaded to apply tone mapping between color channels. We denote this configuration of MuLUT as MuLUT-SDY-X2-C, where ``-C'' denotes MuLUT with channel indexing. This multi-stage and multi-branch structure, enabled by the cooperation of multiple LUTs, addresses the above obstacles of adapting LUT to the task of image demosaicing, showing the versatility of MuLUT.

\subsection{Denoising and Deblocking}

For grayscale image denoising, the pipeline of MuLUT is similar to super-resolution. The difference is that the $\mathtt{pixelshuffle}$ operations in all MuLUT blocks are removed, since there is no need to change the resolution of the input image. This pipeline can be adapted to other grayscale image processing tasks, \emph{e.g.}, grayscale image deblocking, where the spatial resolution is not changed after processing. We apply a configuration of MuLUT with two cascaded stages, where each stage contains all proposed indexing patterns (``S'', ``D'', ``Y'', ``E'', ``H'', and ``O''). We denote this configuration as MuLUT-SDYEHO-X2.


For color image denoising, we adopt a pipeline like the one illustrated in Fig.~\ref{fig:overview}(a), where a channel-wise MuLUT block is inserted between cascaded stages. The three color channels are processed with different spatial-wise LUTs and then concatenated together. After processed across channels with channel-wise LUTs, the image can be further processed by another group of spatial-wise LUTs. Finally, the restored image is obtained by concatenating three color channels again. This pipeline can be adapted to other color image processing tasks as well as video processing tasks. Following the configuration used in grayscale image denoising, we denote the configuration of MuLUT for color image denoising as MuLUT-SDYEHO-X2-C.

%% file: pics/fig5_gmblock.tex
\begin{figure}[t]
    \centering
    \includegraphics[width=0.7\columnwidth]{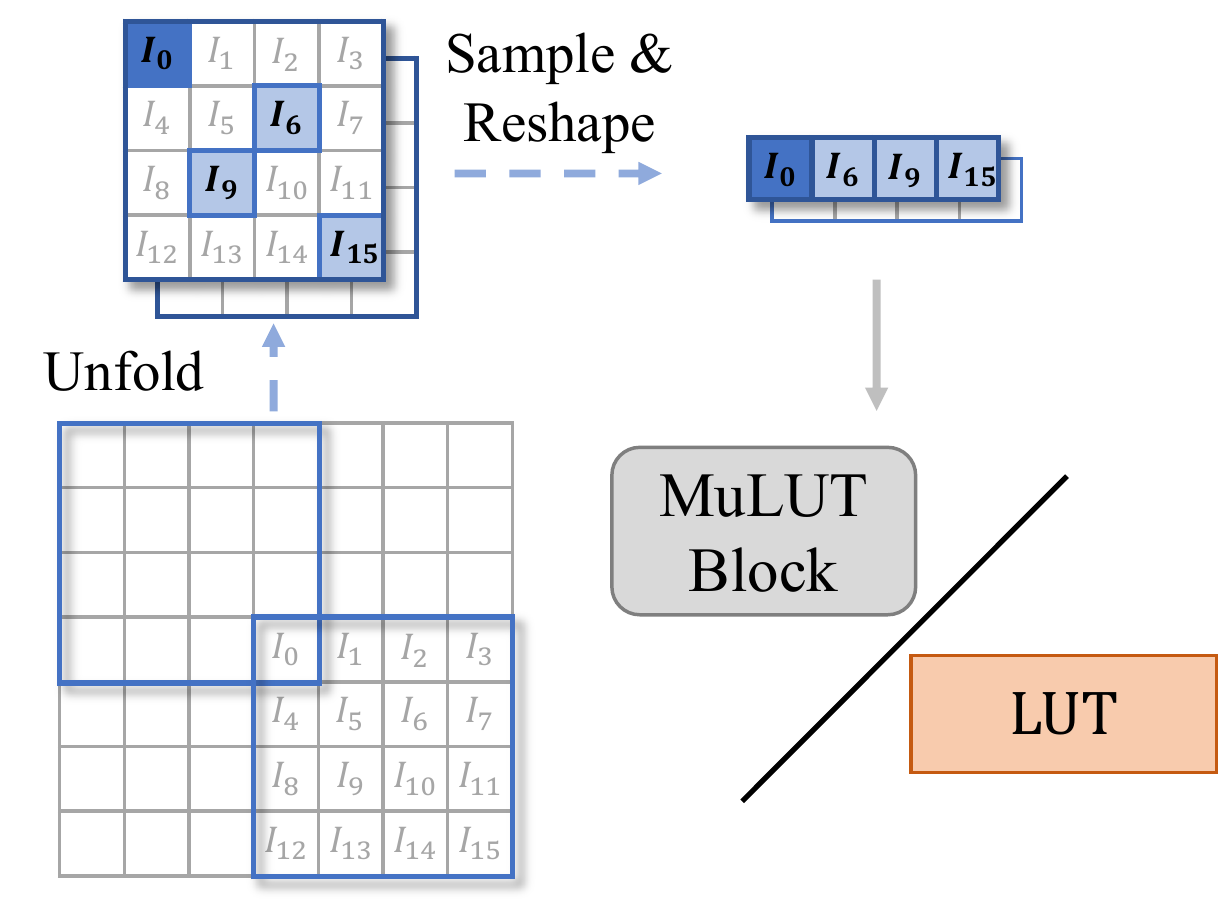}
    \caption{The general implementation of obtaining arbitrary patterns.} 
    \label{fig:gmblock}
\end{figure}

%% file: pics/fig6_hindex.tex
\begin{figure*}[t]
    \begin{center}
    \subfigure[Hierarchical Indexing]{
        \includegraphics[scale=0.5]{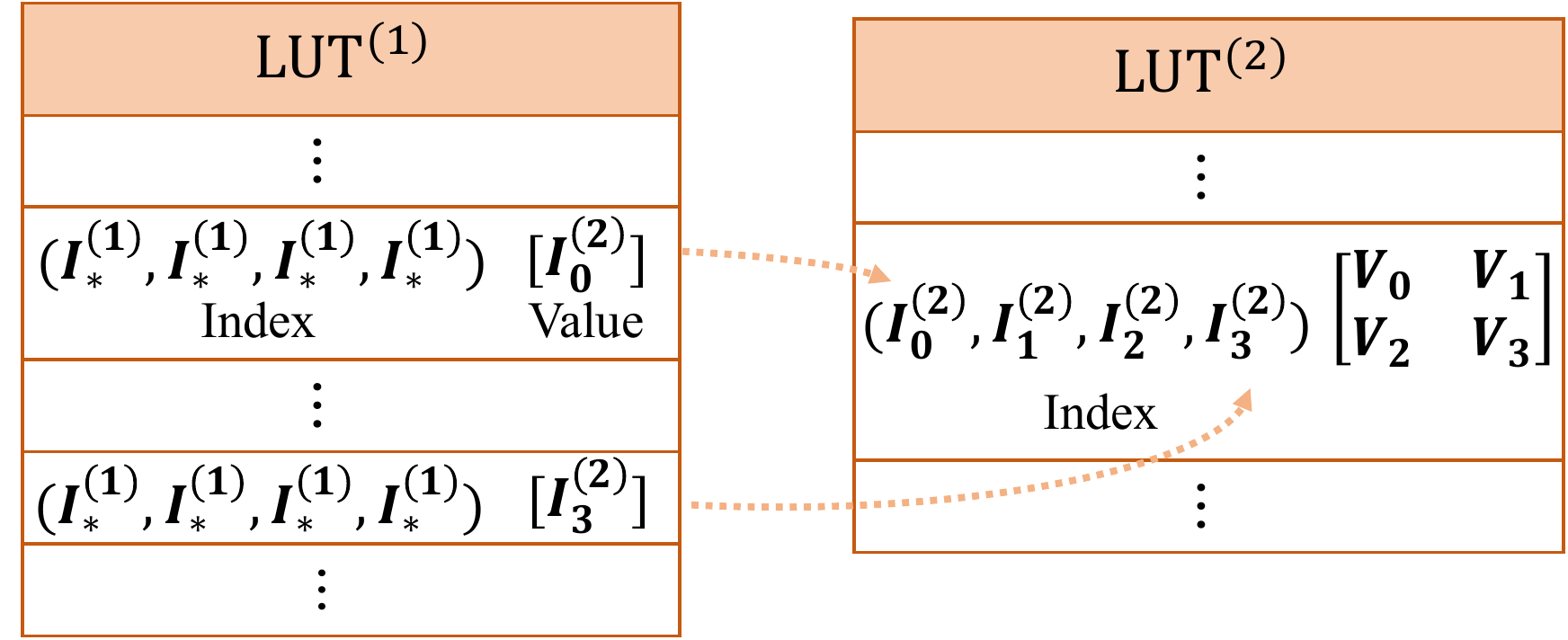}}
    \subfigure[LUT Re-indexing Mechanism]{
        \includegraphics[scale=0.5]{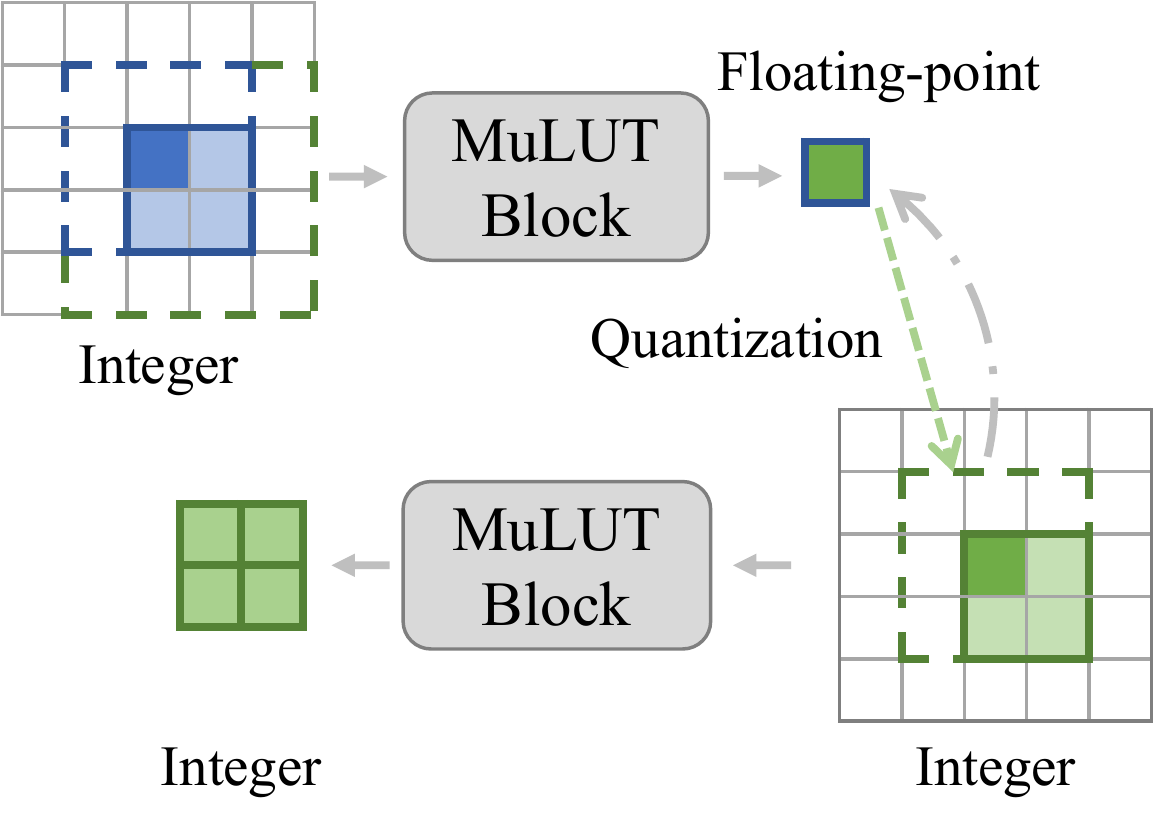}}
    \end{center}
    \caption{(a) The values saved in cascaded LUTs are indexed in a hierarchical way. (b) With LUT re-indexing, the behavior of LUT retrieval is involved in the learning process of the network. Thus, the cascaded LUTs are able to reproduce its performance.} 
    \label{fig:hierarch_index}
\end{figure*}

%% file: pics/fig8_dm.tex
\begin{figure*}[t]
    \begin{center}
        \subfigure[SR-LUT Baseline-A]{
        \includegraphics[width=0.45\textwidth]{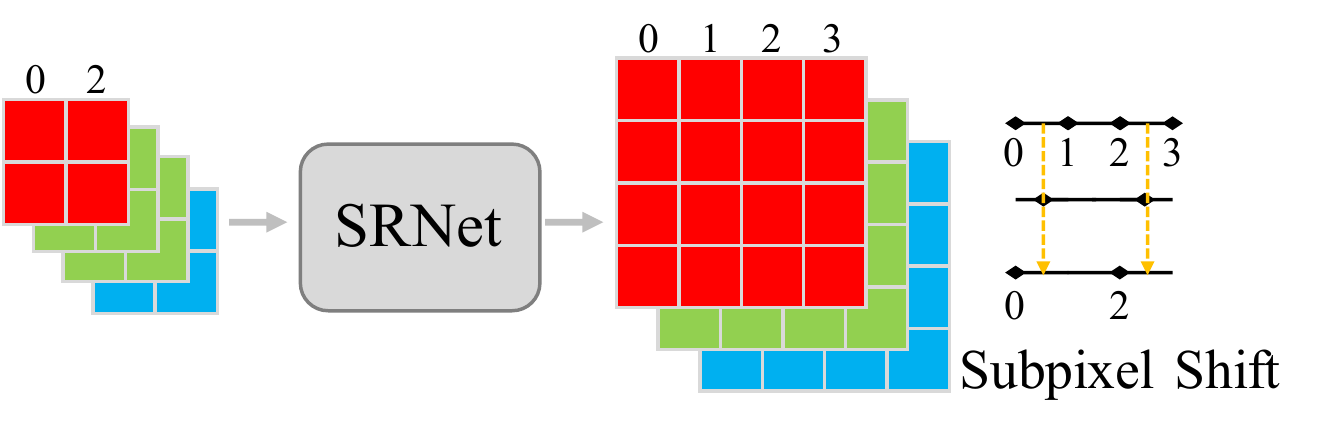}}
        \subfigure[SR-LUT Baseline-B]{
        \includegraphics[width=0.35\textwidth]{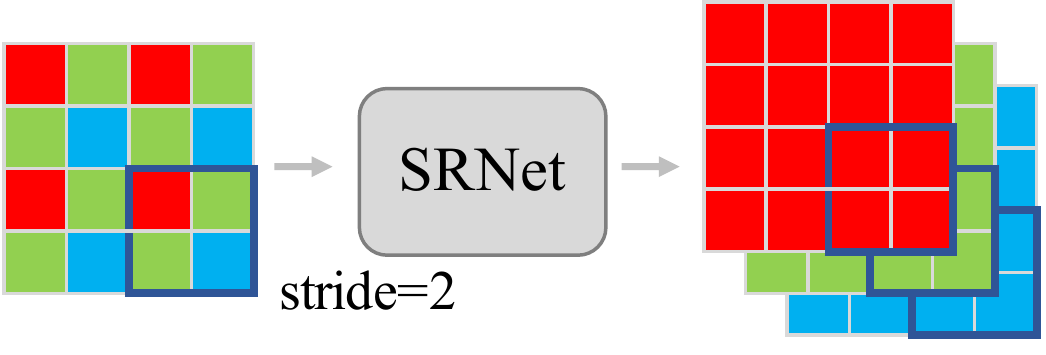}}
        \subfigure[MuLUT]{
        \includegraphics[width=0.8\textwidth]{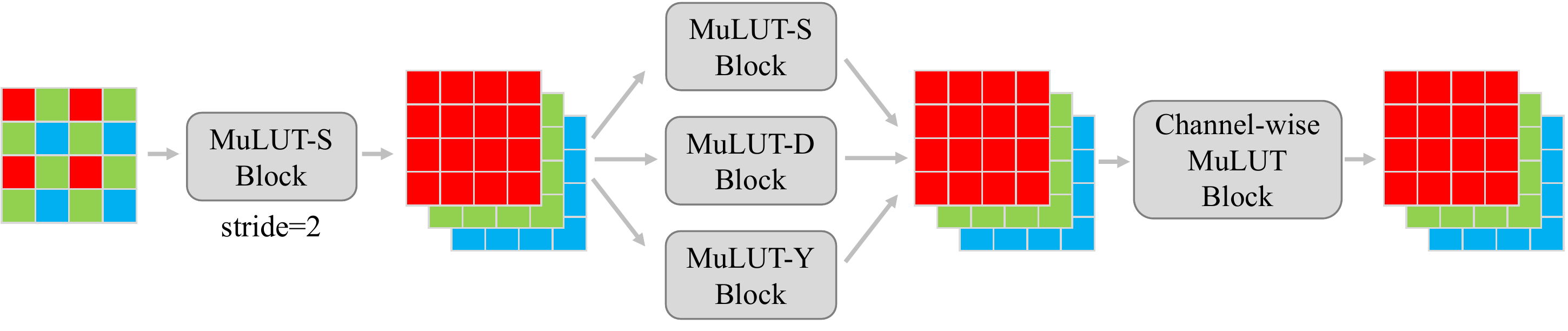}}
    \end{center}
\caption{MuLUT and SR-LUT baselines for image demosaicing. The cooperation of multiple LUTs enables the flexible design of the processing pipeline.}
\label{fig:dm}
\end{figure*}

%% file: pics/fig7_mulut_sr.tex
\begin{figure}[t]
    \centering
    \includegraphics[width=\columnwidth]{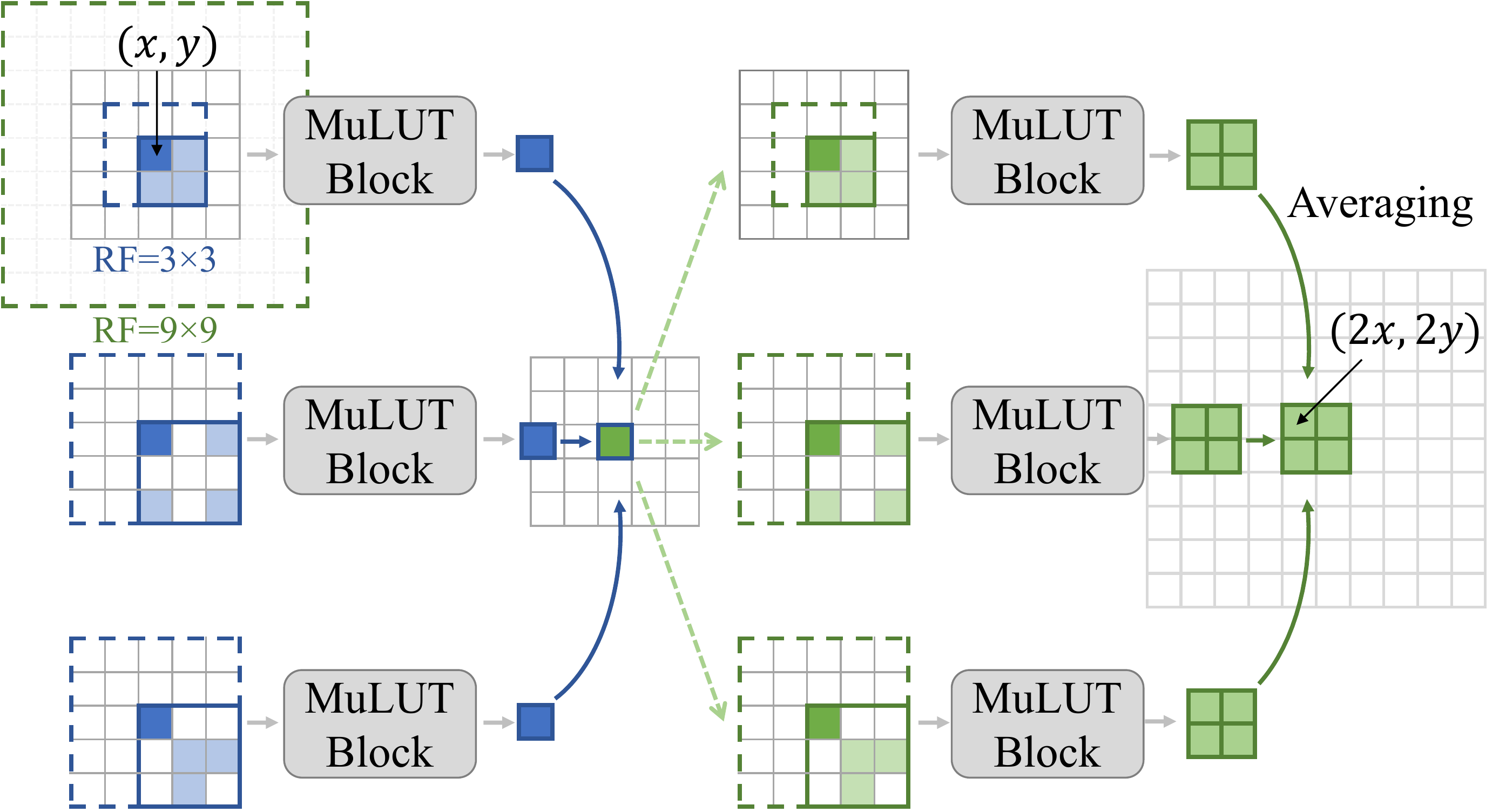}
    \caption{The framework of MuLUT for image super-resolution. With the default MuLUT-SDY-X2 configuration, the RF can be effectively enlarged from $3 \times 3$ to $9 \times 9$.} 
    \label{fig:mulut_sr}
\end{figure}

%% file: tables/config_table.tex
\begin{table}[t]
    \centering
    \caption{Different configurations of MuLUT for representative restoration tasks. }
    \label{tab:config}
    \begin{tabular}{lccc}
    \toprule
    \multicolumn{1}{c}{\begin{tabular}[c]{@{}c@{}} Task \end{tabular}}
     & \multicolumn{1}{c}{\begin{tabular}[c]{@{}c@{}} complementary \\ indexing \end{tabular}} 
     & \multicolumn{1}{c}{\begin{tabular}[c]{@{}c@{}} hierarchical \\ indexing \end{tabular}} 
     & \multicolumn{1}{c}{\begin{tabular}[c]{@{}c@{}} channel \\ indexing \end{tabular}} 
      \\ \midrule
    Super-Resolution       & SDY    & $ \checkmark $    &       \\
    Demosaicing       & SDY    & $\checkmark$    & $\checkmark$    \\
    Grayscale Denoising       & SDYEHO    & $\checkmark$    &     \\
    Color Denoising       & SDYEHO    & $\checkmark$    & $\checkmark$   \\
    Grayscale Deblocking       & SDYEHO    & $\checkmark$    &
    \\ \bottomrule
    \end{tabular}
    \begin{tablenotes}
        \item[1] Note that only the configuration for super-resolution appeared in our previous work \cite{mulut_eccv}, and other configurations are new in this paper.
    \end{tablenotes}
  \end{table}

%% file: tables/energy_table.tex
\begin{table}[t]
    \centering
    \caption{The energy cost of operation in different data types. }
    \begin{tabular}{lrrrr}
    \toprule
    Operation           & int8 & int32 & float16 & float32 \\ \midrule
    Add. (pJ)       & 0.03  & 0.1    & 0.4   & 0.9   \\
    Mult. (pJ) & 0.2   & 3.1    & 1.1   & 3.7  \\ \bottomrule
    \end{tabular}
    \begin{tablenotes}
        \item[1] The numbers are reported in the literature \cite{Dally15,DBLP:conf/isscc/Horowitz14,DBLP:journals/pieee/SzeCYE17}.
    \end{tablenotes}
    \label{tab:energy}
    \end{table}

%% file: tables/comp_table.tex
\begin{table*}[t]  \footnotesize
    \centering
    \caption{The comparison of energy cost and performance of super-resolution. Our method shows superior performance (0.6$\sim$0.8dB gain) over SR-LUT, and a clear advantage (about $100 \times$ less) in terms of energy cost compared with DNN methods, even with their AdderNet and quantized versions.}
    \label{tab:comp}
    \resizebox{0.8\textwidth}{!}{%
    \begin{tabular}{lrrrrrrr|ccc}
    \toprule
    & \multicolumn{1}{c}{\begin{tabular}[c]{@{}c@{}}int8 \\ Add.\end{tabular}} & \multicolumn{1}{c}{\begin{tabular}[c]{@{}c@{}}int8 \\ Mul.\end{tabular}} & \multicolumn{1}{c}{\begin{tabular}[c]{@{}c@{}}int32 \\ Add.\end{tabular}} & \multicolumn{1}{c}{\begin{tabular}[c]{@{}c@{}}int32 \\ Mul.\end{tabular}} & \multicolumn{1}{c}{\begin{tabular}[c]{@{}c@{}}float32 \\ Add.\end{tabular}} & \multicolumn{1}{c}{\begin{tabular}[c]{@{}c@{}}float32 \\ Mul.\end{tabular}} & \multicolumn{1}{c|}{\begin{tabular}[c]{@{}c@{}}Energy \\ Cost(pJ)\end{tabular}} & \multicolumn{1}{c}{\begin{tabular}[c]{@{}c@{}}~~Set14~~~ \\ PSNR \end{tabular}}& \multicolumn{1}{c}{\begin{tabular}[c]{@{}c@{}}BSDS100 \\ PSNR \end{tabular}} & \multicolumn{1}{c}{\begin{tabular}[c]{@{}c@{}}Urban100 \\ PSNR \end{tabular}} \\ \midrule
    Bilinear            &          &           &            &           & 7.4M        & 7.4M         & \cellcolor[HTML]{EFEFEF}33.9M       & 29.15 & 28.65   & 25.95    \\ 
    Bicubic            &          &           &            &            & 14.7M        & 14.7M        & \cellcolor[HTML]{EFEFEF}67.8M       & 30.23 & 29.53   & 26.86    \\ \midrule
    SR-LUT (F variant) \cite{DBLP:conf/cvpr/JoK21}          & 13.6M    & 0.5M      & 11.8M      & 19.1M       &             &             &\cellcolor[HTML]{EFEFEF}61.0M      & 31.88 & 30.77   & 28.49    \\
    SR-LUT \cite{DBLP:conf/cvpr/JoK21}          &  19.1M    & 0.5M      & 28.6M      & 22.8M       &             &             & \cellcolor[HTML]{EFEFEF}74.2M       & 31.73 & 30.64   & 28.50    \\
    MuLUT-SDY       & 56.9M    & 0.5M      & 118.0M      & 68.0M        &             &             & \cellcolor[HTML]{EFEFEF}224.3M    & 32.35 & 31.17   & 29.10    \\
    MuLUT-SDY-X2 & 80.6M    & 0.9M      & 109.4M     & 85.5M       &             &             & \cellcolor[HTML]{EFEFEF}278.5M      & \textbf{32.49} & \textbf{31.23}   & \textbf{29.31}    \\ 
    \midrule
    FSRCNN \cite{DBLP:conf/eccv/DongLT16}           &          &           &            &            & 6.1G        & 6.1G         & \cellcolor[HTML]{EFEFEF}28.1G       & 32.69 & 31.49   & 29.87    \\ 
    A-VDSR-8bit \cite{DBLP:conf/cvpr/Song0C0XT21}       & 1224.1G  & 1.1G       &            &            &             &             & \cellcolor[HTML]{EFEFEF}36.9G       & 32.85 & 31.66   & 30.07    \\
    A-VDSR \cite{DBLP:conf/cvpr/Song0C0XT21}            &          &           &            &            & 1224.1G     & 1.1G         & \cellcolor[HTML]{EFEFEF}1105.6G     & 32.93 & 31.81   & 30.48    \\
    VDSR \cite{DBLP:conf/cvpr/KimLL16a}             &          &           &            &            & 612.6G      & 612.6G       & \cellcolor[HTML]{EFEFEF}2817.9G     & 33.03 & 31.90   & 30.76    \\ 
    A-CARN-1/4 \cite{DBLP:conf/cvpr/Song0C0XT21}        &          &           &            &            & 28.9G       & 0.1G         & \cellcolor[HTML]{EFEFEF}26.3G       & -     &  -      & 30.21    \\
    CARN-1/4 \cite{DBLP:conf/cvpr/Song0C0XT21}          &          &           &            &            & 14.5G       & 14.5G        & \cellcolor[HTML]{EFEFEF}66.5G       & -     &  -      & 30.40    \\ 
    CARN-M \cite{DBLP:conf/eccv/AhnKS18}           &          &           &            &            & 91.2G       & 91.2G        & \cellcolor[HTML]{EFEFEF}419.5G      & 33.17 & 31.88   & 31.23    \\ 
    RCAN \cite{DBLP:conf/eccv/ZhangLLWZF18}           &          &           &            &            & 3.53T       & 3.53T        & \cellcolor[HTML]{EFEFEF}16.2T       & 34.14     &  32.41      & 33.17    \\ 
    SwinIR \cite{DBLP:conf/iccvw/LiangCSZGT21}          &          &           &            &            & 195.6G       & 195.6G        & \cellcolor[HTML]{EFEFEF}899.8G       & 33.86     &  32.31      & 32.76    \\ 
    \bottomrule
    \end{tabular}%
    }
    \begin{tablenotes}
        \item[1] The energy cost and performance are evaluated for producing a $1280 \times 720$ HD image through $2\times$ super-resolution. 
        \item[2] The statistics of operations not involved in a method are left blank. 
        \item[3] A-VDSR denotes the AdderNet version of VDSR \cite{DBLP:conf/cvpr/KimLL16a,DBLP:conf/cvpr/Song0C0XT21}. A-VDSR-8bit denotes performing 8bit quantization for A-VDSR. CARN-1/4 is a lightweight version of CARN \cite{DBLP:conf/eccv/AhnKS18,DBLP:conf/cvpr/Song0C0XT21}, and A-CARN-1/4 denotes its AdderNet version.
    \end{tablenotes}
    \end{table*}

%% file: pics/fig9_sr_tradeoff.tex
\begin{figure}[t]
    \centering
    \includegraphics[width=0.9\columnwidth]{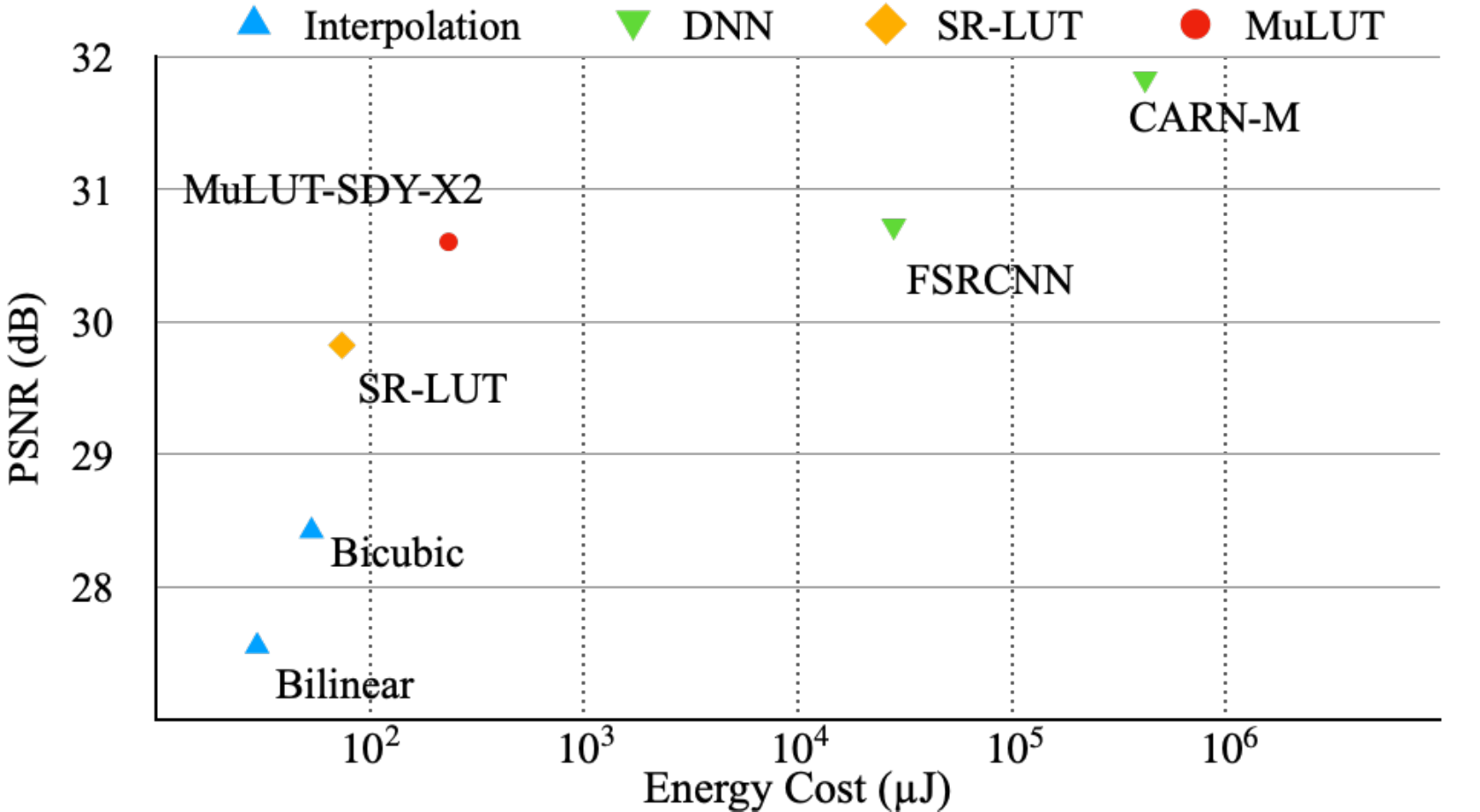}
    \caption{MuLUT achieves a better performance and efficiency tradeoff, compared with interpolation, SR-LUT, and lightweight DNN methods. The PSNR values are evaluated on Set5 for $4 \times$ SR.} 
    \label{fig:sr_tradeoff}
\end{figure}

%% file: parts/experiments.tex
\section{Experiments and Results}

\subsection{Implementation Details}
We train the MuLUTNet on the DIV2K dataset \cite{DBLP:conf/cvpr/AgustssonT17}, which is widely used in image restoration tasks. The DIV2K dataset contains 800 training images and 100 validation images with 2K resolution. It covers multiple scenes and encapsulates diverse patches. We train the MuLUTNet with the Adam optimizer \cite{DBLP:journals/corr/KingmaB14} in the cosine annealing schedule \cite{DBLP:conf/iclr/LoshchilovH17} at a learning rate of $1 \times 10^{-4}$. We use the mean-squared error (MSE) loss function as the optimization target. The MuLUTNet is trained for $2 \times 10^5$ iterations at a batch size of $32$ and at a patch size of $48$. The cached LUTs are uniformly sampled with interval $2^4$, \emph{i.e.}, from LUT[0-255] to LUT[0-16]. After locating coordinates, the final prediction is obtained with 4D simplex interpolation \cite{DBLP:conf/cvpr/JoK21}, a 4D equivalent of 3D tetrahedral interpolation \cite{DBLP:journals/jei/KassonNPH95}. We further finetune the cached LUTs on the same training dataset for 2,000 iterations with the proposed LUT-aware finetuning strategy. Our method is implemented in the PyTorch library \cite{DBLP:conf/nips/PaszkeGMLBCKLGA19}, and the experiments are conducted on 2 NVIDIA GTX 3090 GPUs. 



\subsection{Efficiency Evaluation}

We conduct efficiency evaluation for different kinds of methods in terms of theoretical energy cost, real-world running time, and storage occupation.

\noindent\textbf{Energy cost.}
Taking single-image super-resolution as an example, we estimate the theoretical energy cost of interpolation, LUT-based, and DNN methods, following the protocol in AdderSR \cite{DBLP:conf/cvpr/Song0C0XT21}. We illustrate the tradeoff between restoration performance (evaluated with PSNR) and energy cost in Fig.~\ref{fig:sr_tradeoff}. As can be seen, with similar energy cost as interpolation and SR-LUT, MuLUT obtains comparable restoration performance to lightweight DNN methods (\emph{e.g.} FSRCNN \cite{DBLP:conf/eccv/DongLT16} and CARN-M \cite{DBLP:conf/eccv/AhnKS18}), achieving a better performance and efficiency tradeoff. We further compare the energy cost with the AdderNet \cite{DBLP:conf/cvpr/Song0C0XT21} and quantized versions of VDSR \cite{DBLP:conf/cvpr/KimLL16a} and CARN \cite{DBLP:conf/eccv/AhnKS18}. We calculate the statistics of multiplications and additions in different data types required by each method and estimate their total energy cost. Our estimation is based on Table~\ref{tab:energy}, where the theoretical energy cost for each operation in different data types is reported. The detailed comparison is listed in Table~\ref{tab:comp}. As can be seen, our method shows superior performance compared with interpolation methods and SR-LUT. For example, MuLUT-SDY-X2 exceeds SR-LUT by 0.6$\sim$0.8dB, while maintaining a similar energy cost. On the other hand, MuLUT maintains a clear energy cost advantage over DNN methods, even their AdderNet and quantized versions. Compared with FSRCNN, A-VDSR-8-bit, and A-CARN-1/4, MuLUT costs about $100 \times$ less energy while achieving comparable restoration performance. 


\input{tables/runtime_table_sr.tex}

\input{tables/runtime_table_dn.tex}

\noindent\textbf{Running time.}
Besides, following SR-LUT \cite{DBLP:conf/cvpr/JoK21}, we implement the proposed MuLUT on the widely used ANDROID platform and report the running times of interpolation, LUT-based, sparse coding, and DNN methods in Table~\ref{tab:runtime_sr}. As listed, MuLUT maintains the efficiency of SR-LUT, showing a clear advantage compared to sparse coding methods and DNN methods. Note that the CPU computing architecture is not optimized for LUT, which can be embedded into on-device memory, such as those of image processors in consumer cameras for low-latency execution \cite{DBLP:conf/dac/DengZZY19,9380930}. Moreover, MuLUT can be implemented without modern computing libraries like PyTorch, thus having better practicality on edge devices. For another task, image denoising, we also report the running times of LUT-based, classical, and DNN methods in Table~\ref{tab:runtime_dn}. As can be seen, while LUT-based methods show a clear advantage over classical and DNN methods, MuLUT outperforms SR-LUT significantly in PSNR at a linearly growing cost in running time.

\input{tables/main_table_sr.tex}

\noindent\textbf{Storage occupation.}
Finally, we analyze the extra storage required by different kinds of methods. For LUT-based and sparse coding methods, the storage size is measured by their dictionary and table size. For DNN methods, we report the number of parameters. As listed in Table~\ref{tab:runtime_sr} and Table~\ref{tab:runtime_dn}, the size of MuLUT is similar to that of SR-LUT. Thus, it also can be cached into onboard memory for highly efficient access. In addition, as mentioned above, MuLUT requires simpler dependencies compared to DNN methods, leading to fewer storage requirements for execution libraries.

In summary, MuLUT shows its advantage in efficiency and practicality for real-world deployment on edge devices compared to DNN methods.

\subsection{Performance Evaluation} 

\subsubsection{Image Super-Resolution}

\noindent\textbf{Datasets and metrics.}
For image super-resolution, we evaluate our method on five benchmark datasets: Set5, Set14, BSDS100 \cite{DBLP:conf/iccv/MartinFTM01}, Urban100 \cite{DBLP:conf/cvpr/HuangSA15}, and Manga109 \cite{DBLP:journals/mta/MatsuiIAFOYA17}. For quantitative evaluation, we report peak signal-to-noise ratio (PSNR) and structural similarity index (SSIM) \cite{DBLP:journals/tip/WangBSS04}, which are widely used for image quality assessment in terms of restoration fidelity.

\noindent\textbf{Comparison methods.}
We compare our method with various single-image super-resolution methods, including interpolation-based methods (nearest neighbor, bilinear, and bicubic interpolation), sparse coding methods (NE + LLE \cite{DBLP:conf/cvpr/ChangYX04}, Zeyde et al. \cite{DBLP:conf/cas/ZeydeEP10}, ANR \cite{DBLP:conf/iccv/TimofteDG13}, and A+ \cite{DBLP:conf/accv/TimofteSG14}), SR-LUT \cite{DBLP:conf/cvpr/JoK21}, and DNN methods (FSRCNN \cite{DBLP:conf/eccv/DongLT16}, CARN-M \cite{DBLP:conf/eccv/AhnKS18}, RCAN \cite{DBLP:conf/eccv/ZhangLLWZF18}, and SwinIR \cite{DBLP:conf/iccvw/LiangCSZGT21}). 

\noindent\textbf{Quantitative evaluation.}
The quantitative comparisons of different methods for $\times 2$, $\times 3$, and $\times 4$ super-resolution are listed in Table~\ref{tab:main_sr}. Overall, our method obtains comparable performance with FSRCNN while boosting the performance over SR-LUT significantly. For example, with 2 cascaded stages and 3 parallel blocks, MuLUT-SDY-X2 enlarges the RF size from $3\times 3$ to $9\times 9$, yields 1.1dB PSNR gain over SR-LUT on the Manga109 dataset ($\times 4$) and even exceeds FSRCNN in terms of SSIM. With only complementary indexing, MuLUT-SDY enlarges the RF from $3 \times 3$ to $5 \times 5$, improving the PSNR value over SR-LUT by 0.72dB on the same dataset. 


\input{pics/visual_sr_main.tex}

\setcounter{figure}{11}    
\input{pics/visual_dm_main.tex}
\setcounter{figure}{10}    
\input{pics/fig_dm_tradeoff.tex}
\setcounter{figure}{12}    

\setcounter{figure}{13}    
\input{pics/visual_dn_main.tex}
\input{pics/visual_cdn_main.tex}

\noindent\textbf{Qualitative evaluation.}
We compare the visual quality of our method (MuLUT-SDY-X2) with other methods in Fig.~\ref{fig:visual_sr_main}. In the first three examples, SR-LUT produces notable artifacts, \emph{e.g.}, along the border of the wing (\textit{butterfly} form Set5). MuLUT-SDY-X2 achieves similar visual quality to A+ and FSRCNN. In the last three examples, our method is able to generate sharper edges and obtain better visual quality than A+ and FSRCNN, \emph{e.g.}, the eyebrow of the character (\textit{TetsuSan} from Manga109). To sum up, our method achieves better visual quality than SR-LUT and comparable visual quality with A+ and FSRCNN.

\subsubsection{Image Demosaicing}

\noindent\textbf{Datasets and metrics.}
We sample pixels according to the Bayer pattern to construct synthetic data pairs for image demosaicing, where the mosaiced images are simulated by applying color masks on the original images. We evaluate our method on the widely-used Kodak \cite{Li2008ImageDA} and McMaster \cite{DBLP:journals/jei/0006WB011} datasets. We report the cPSNR metric, which is averaged across three color channels.


\noindent\textbf{Comparison methods.}
Besides the two single-LUT baselines illustrated in Fig.~\ref{fig:dm}, we compare our method with bilinear interpolation, a classical method (D-LMMSE \cite{DBLP:journals/tip/ZhangW05}), and a DNN method (DemosaicNet \cite{DBLP:journals/tog/Durand16a}).

\noindent\textbf{Quantitative evaluation.}
As illustrated in Fig.~\ref{fig:dm_tradeoff}, MuLUT-SDY-X2 and MuLUT-SDY-X2-C improve the performance of single-LUT baselines by a large margin, \emph{e.g.}, over 6.0dB on the Kodak dataset, while achieving a better performance and efficiency tradeoff compared with D-LMMSE and DemosaicNet. 

\noindent\textbf{Qualitative evaluation.}
As shown in Fig.~\ref{fig:visual_dm_main}, the results of Baseline-A are blurry because of subpixel shift, and Baseline-B produces noticeable blocking artifacts due to limited RF, while MuLUT-SDY-X2-C obtains comparable visual quality with computation-heavy D-LMMSE and DemosaicNet.

\input{tables/main_table_dn.tex}

\setcounter{figure}{12}

\input{pics/fig_cdn_tradeoff.tex}
\setcounter{figure}{15}


\input{pics/visual_db_main.tex}

\input{tables/main_table_db.tex}

\subsubsection{Image Denoising}

\noindent\textbf{Datasets and metrics.}
For grayscale image denoising, we evaluate our method with two benchmark datasets: Set12 and BSD68. For color image denoising, we evaluate our method on three datasets: CBSD68, Kodak24, and McMaster. For quantitative evaluation, we report PSNR for grayscale image denoising, and we report cPSNR for color image denoising.

\noindent\textbf{Comparison methods.}
We compare our method with various single-image denoising methods, including single-LUT solution (SR-LUT \cite{DBLP:conf/cvpr/JoK21}), classical methods (BM3D \cite{DBLP:journals/tip/DabovFKE07}, WNNM \cite{DBLP:conf/cvpr/GuZZF14}, and TNRD \cite{DBLP:journals/pami/ChenP17}), and DNN methods (DnCNN \cite{DBLP:journals/tip/ZhangZCM017}, FFDNet \cite{DBLP:journals/tip/ZhangZZ18}, and SwinIR \cite{DBLP:conf/iccvw/LiangCSZGT21}). We adapt SR-LUT to denoising as the single-LUT baseline by removing its $\mathtt{pixelshuffle}$ operation and retraining it on noisy and clean data pairs.

\noindent\textbf{Quantitative evaluation.}
The quantitative comparisons with other methods for grayscale image denoising at different noise levels are listed in Table~\ref{tab:main_dn}. As can be seen, MuLUT-SDY-X2 ($9 \times 9$ RF) and MuLUT-SDYEHO-X2 ($13 \times 13$ RF) improve the restoration performance over SR-LUT ($3 \times 3$ RF) significantly, showing that the RF size has a crucial influence on the performance of denoising methods. Besides, MuLUT-SDYEHO-X2 approaches a comparable performance with BM3D, \emph{e.g.}, 28.34dB vs 28.57dB on the BSD68 dataset at a noise level of 25. For color image denoising, we illustrate the performance and running time tradeoff in Fig.~\ref{fig:cdn_tradeoff}. As can be seen, MuLUT-SDY-X2 and MuLUT-SDYEHO-X2-C outperform SR-LUT by a large margin, \emph{e.g.}, 3.0dB and 3.8dB gain over single-LUT baseline, respectively, on the Kodak24 dataset at a noise level of 50, while achieving a better performance and efficiency tradeoff compared with CBM3D, DnCNN, and SwinIR. We further investigate the effectiveness of larger RF and channel indexing in Table~\ref{tab:c_cdn_abl}.



\noindent\textbf{Qualitative evaluation.}
For grayscale image denoising, we show qualitative comparisons in Fig.~\ref{fig:visual_dn_main}. As can be seen, the results of MuLUT-SDYEHO-X2 are cleaner than those of SR-LUT. With a limited RF, SR-LUT fails to distinguish the noise and signal, resulting in corrupted structures and noisy outputs. MuLUT-SDYEHO-X2 is able to obtain similar visual quality to BM3D and DnCNN. For color image denoising, we show qualitative comparisons in Fig.~\ref{fig:visual_cdn_main}. As can be seen, MuLUT-SDYEHO-X2-C produces cleaner edges than SR-LUT. Moreover, the textures generated by MuLUT-SDYEHO-X2-C are finer than those by CBM3D, since CBM3D tends to produce blurry results where the matching patches are hard to find.

\input{tables/abl_table_net.tex}

\subsubsection{Image Deblocking}

\noindent\textbf{Datasets and metrics.}
For image deblocking, we evaluate our method with two widely used benchmark datasets: Classic5 and LIVE1. For quantitative evaluation, besides PSNR and SSIM, we also report the PSNR-B metric, which is designed to evaluate the blocking effects in images.

\noindent\textbf{Comparison methods.}
We compare our method with various image deblocking methods, including single-LUT solution (SR-LUT \cite{DBLP:conf/cvpr/JoK21}), classical method (SA-DCT \cite{DBLP:journals/tip/FoiKE07}), and DNN methods (ARCNN \cite{DBLP:conf/iccv/DongDLT15} and SwinIR \cite{DBLP:conf/iccvw/LiangCSZGT21}). The single-LUT baseline is obtained similarly to that in denoising.

\noindent\textbf{Quantitative evaluation.}
The quantitative comparisons with other methods for image deblocking at different quality factors are listed in Table~\ref{tab:main_db}. As can be seen, MuLUT-SDY-X2 and MuLUT-SDYEHO-X2 outperform SR-LUT and SA-DCT, especially in terms of PSNR-B. For example, on the LIVE1 dataset at a quality factor of 30, MuLUT-SDY-X2 and MuLUT-SDYEHO-X2 improve SR-LUT by 0.80dB and 0.85dB PSNR-B gain, respectively. Besides, with a larger RF, MuLUT-SDYEHO-X2 outperforms MuLUT-SDY-X2, approaching comparable performance with ARCNN, \emph{e.g.}, 32.84dB vs 33.14dB in terms of PSNR-B on the LIVE1 dataset at a quality factor of 40.

\noindent\textbf{Qualitative evaluation.}
We compare the visual quality of MuLUT-SDYEHO-X2 with other methods in Fig.~\ref{fig:visual_db_main}. As can be seen, the results of SR-LUT contain noticeable blocking artifacts, SA-DCT sometimes loses details and produces over-smooth results (the figures in the third example), while the results of MuLUT-SDYEHO-X2 show similar visual quality to those of ARCNN.

\subsection{Ablation Analysis}

We conduct several ablation experiments to verify the effectiveness of the design principles of MuLUT.

\noindent\textbf{Analysis of the network capacity.} 
We conduct experiments to investigate the influence of network capacity of a MuLUT block. As listed in Table~\ref{tab:net_abl}, for both SR-LUT and MuLUT, the number of filters has a limited influence on the performance. On the other hand, with a similar number of parameters, MuLUT-SDY-X2 (nf.=64) with an RF size of $9 \times 9$ outperforms SR-LUT (nf.=256), showing the critical role of RF size. Besides, the dense connection not only helps the convergence but also boosts the performance.

\input{tables/abl_table_patterns.tex}
\input{tables/abl_table_hierarchical.tex}

\noindent\textbf{The effectiveness of complementary indexing.} 
We conduct experiments with combinations of different indexing patterns of parallel LUTs. As listed in Table~\ref{tab:p_abl}, with MuLUT-S and MuLUT-D working together, MuLUT-SD is able to cover a region of $5 \times 5$, but not all pixels are covered. Still, it significantly improves the performance of SR-LUT and outperforms MuLUT-SSS with the repeating indexing patterns. Besides, we include another ``T'' pattern, where $I_0, I_2, I_5, I_8$ are indexed. Interestingly, the performance of MuLUT-SDT is better than MULUT-SD, even with no new pixels covered, indicating that different indexing patterns convey structure clues. Further, involving the novel ``Y'' shape indexing pattern, MuLUT-SDY covers all pixels in a $5 \times 5$ region and improves the performance, showing the effectiveness of complementary indexing. Similar results can also be observed for involving ``E'', ``H'', and ``O'' patterns. 


\noindent\textbf{The effectiveness of hierarchical indexing.} 
We conduct an experiment with cascading different stages of LUTs. As listed in Table~\ref{tab:h_abl}, cascading more stages enlarges the RF steadily, and the performance improves accordingly. Without LUT re-indexing, the performance drops due to the inconsistency between the super-resolution network and the cached LUT. 
Note that cascading LUTs involves \textit{sub-linear} extra computational burden and storage space, since all LUTs except the ones in the last stage cache only one value for each index entry. Furthermore, with both complementary indexing and hierarchical indexing, MuLUT-SDY-X2 achieves better restoration performance over MuLUT-SDY.


\input{tables/abl_table_hybrid_dm.tex}
\input{tables/abl_table_hybrid_cdn.tex}

\noindent\textbf{The effectiveness of channel indexing.} 
We validate the influence of channel indexing by removing the channel-wise MuLUT block in color image processing. As listed in Table~\ref{tab:c_dm_abl} and Table~\ref{tab:c_cdn_abl}, for both image demosaicing and color image denoising, channel indexing improves the performance at a minor additional energy cost, showing the importance of allowing channel interaction for color image processing.

\noindent\textbf{The effectiveness of LUT-aware finetuning.} 
We compare SR-LUT and MuLUT-SDY-X2 with or without the LUT-aware finetuning strategy. We also report the corresponding network performance. As can be seen in Table~\ref{tab:ft_abl}, there is a performance drop from network predictions to LUT results, especially for the one with larger sampling intervals (3bit LUT). The proposed LUT-aware finetuning strategy is able to fill this gap consistently for different sampling intervals. Especially, after finetuning, a 3bit LUT achieves similar performance compared with a 4bit LUT, while taking $10 \times$ less storage. Further, the proposed MuLUT-SDY-X2 also benefits from the finetuning strategy, showing its effectiveness.

\input{tables/abl_table_ft.tex}

%% file: tables/runtime_table_sr.tex
\begin{table}[t]
\renewcommand\arraystretch{1.1}
    \centering
    \caption{Runtime and storage occupation comparison for generating a $1280 \times 720$ HD image through $4 \times$ super-resolution.}
    \resizebox{\columnwidth}{!}{
        \begin{tabular}{llr|rrr}        
        \toprule
                                & Method              & Platform        & RunTime(ms)   &Storage             & PSNR                                                            \\ \midrule
        \multirow{3}{*}{Interpolation}       
& Nearest                       & Mobile & 9          & -           & 23.45                                \\
& Bilinear                     & Mobile & 20          & -              & 24.21                             \\
& Bicubic                        & Mobile & 97        & -             & 24.91                                \\ \midrule
        \multirow{3}{*}{LUT}             
& SR-LUT                  & Mobile & 137         & 1.274MB          & 26.80                                  \\
& MuLUT-SDY              & Mobile & 228            & 3.823MB         & 27.52                                \\
& MuLUT-SDY-X2                 & Mobile & 242        & 4.062MB         & 27.90                                    \\ \midrule
\multirow{4}{*}{Sparse Coding}             
& NE + LLE \cite{DBLP:conf/cvpr/ChangYX04}                  & PC & 4,687          &1.434MB          & 26.10                                 \\
& Zeyde et al. \cite{DBLP:conf/cas/ZeydeEP10}              & PC & 7,786        &1.434MB              & 26.24                                   \\
& ANR \cite{DBLP:conf/iccv/TimofteDG13}                 & PC & 1,260       &1.434MB      & 26.18                                            \\ 
& A+ \cite{DBLP:conf/accv/TimofteSG14} & PC & 1,151 &15.171MB   & 26.91  \\ \midrule
        \multirow{4}{*}{DNN}             
& FSRCNN \cite{DBLP:conf/eccv/DongLT16}                     & Mobile & 350        &12K  & 27.91                                           \\
& CARN-M \cite{DBLP:conf/eccv/AhnKS18}                     & Mobile & 3,300      &412K   & 29.85                                            \\
& RCAN  \cite{DBLP:conf/eccv/ZhangLLWZF18}                    & Mobile & 13,987    &15,592K   & 31.45                                              \\ 
& SwinIR  \cite{DBLP:conf/iccvw/LiangCSZGT21}                    & Mobile & 26,162  &897K     & 30.92                                             \\ 
\bottomrule
        \end{tabular}
    }
    \begin{tablenotes}
        \item[1] We implement MuLUT, interpolation-based methods, and DNN methods on the ANDROID platform and test them on a Xiaomi 11 mobile smartphone with a Qualcomm Snapdragon 888 CPU. The DNN methods are implemented in the CPU version of the PyTorch library \cite{DBLP:conf/nips/PaszkeGMLBCKLGA19}. For SR-LUT, we test the official implementation provided by the authors.
        \item[2] We test sparse coding methods in their official MATLAB implementations on a personal computer with an Intel Core i5-8700K CPU.
        \item[3] For LUT-based and sparse coding methods, the storage size is measured by their dictionary and table size. For DNN methods, we report the number of parameters.
        \item[4] PSNR values are evaluated on the Manga109 Dataset.
    \end{tablenotes}
\label{tab:runtime_sr}
\end{table}

%% file: tables/runtime_table_dn.tex
\begin{table}[t]
\renewcommand\arraystretch{1.1}
    \centering
    \caption{Runtime and storage occupation comparison for denoising grayscale images at different resolutions.}
    \resizebox{\columnwidth}{!}{
        \begin{tabular}{llr|rrrr}        
        \toprule
        & \multicolumn{1}{c}{\begin{tabular}[c]{@{}c@{}} Method \end{tabular}} & \multicolumn{1}{c}{\begin{tabular}[c]{@{}c@{}} Platform \end{tabular}}
         & \multicolumn{1}{c}{\begin{tabular}[c]{@{}c@{}} RunTime(ms) \\$256 \times 256$ \end{tabular}} 
         & \multicolumn{1}{c}{\begin{tabular}[c]{@{}c@{}} RunTime(ms) \\ $512 \times 512$\end{tabular}} 
         & \multicolumn{1}{c}{\begin{tabular}[c]{@{}c@{}} Storage \end{tabular}} 
         & \multicolumn{1}{c}{\begin{tabular}[c]{@{}c@{}} PSNR \end{tabular}}  \\ \midrule
        \multirow{3}{*}{LUT}             
& SR-LUT \cite{DBLP:conf/cvpr/JoK21}                  & Mobile & 7  & 21   & 82KB    & 26.85                                             \\
& MuLUT-SDY-X2                 & Mobile &  26  & 99  & 489KB    & 28.18                                             \\ 
& MuLUT-SDYEHO-X2                 & Mobile &  51  & 195  & 978KB    & 28.34                                             \\ \midrule
\multirow{3}{*}{Classical}             
& BM3D \cite{DBLP:journals/tip/DabovFKE07}                 & PC & 2,599 & 12,481     & -   & 28.57                                            \\
& WNNM \cite{DBLP:conf/cvpr/GuZZF14}              & PC & 84,734 & 352,732       & - & 28.83                                              \\
& TNRD  \cite{DBLP:journals/pami/ChenP17}                & PC & 1,140 & 1,564     & -  & 28.92                                            \\ 
\midrule
        \multirow{3}{*}{DNN}             
& DnCNN \cite{DBLP:journals/tip/ZhangZCM017}                    & Mobile & 635   & 2,497     &555K   & 29.23                                            \\
& FFDNet \cite{DBLP:journals/tip/ZhangZZ18}                    & Mobile & 167   & 550  &485K   & 29.19                                              \\
& SwinIR \cite{DBLP:conf/iccvw/LiangCSZGT21}                    & Mobile & 94,849   & 362,082  & 11,449K  & 29.50                                             \\ 
\bottomrule
        \end{tabular}
    }
\begin{tablenotes}
    \item[1] The evaluation environments are the same as SR. Classical methods are tested with their official MATLAB implementation.
    \item[2] For LUT-based, the storage size is measured by their table size. For DNN methods, we report the number of parameters.
    \item[3] PSNR values are evaluated on the BSD68 Dataset.
\end{tablenotes}
\label{tab:runtime_dn}
\end{table}

%% file: tables/main_table_sr.tex
\begin{table*}[t]  \footnotesize
    \caption{The comparison with other methods for image super-resolution on standard benchmark datasets. With enlarged RF, MuLUT achieves a significant improvement in restoration performance over SR-LUT.}
    \label{tab:main_sr}
    \centering
    \resizebox{\textwidth}{!}{%
     \begin{tabular}{llcccccccccccc}
         \toprule
     \multirow{2}{*}{}                & \multirow{2}{*}{Method}  & \multirow{2}{*}{Scale} & \multirow{2}{*}{RF Size} & \multicolumn{2}{c}{Set5} & \multicolumn{2}{c}{Set14} & \multicolumn{2}{c}{BSDS100} & \multicolumn{2}{c}{Urban100} & \multicolumn{2}{c}{Manga109} \\ \cmidrule(lr){5-6} \cmidrule(lr){7-8} \cmidrule(lr){9-10} \cmidrule(lr){11-12} \cmidrule(lr){13-14} 
     &    & &                   & PSNR       & SSIM        & PSNR        & SSIM        & PSNR         & SSIM         & PSNR         & SSIM          & PSNR         & SSIM          \\
     \midrule
     \multirow{3}{*}{Interpolation}   & Nearest     & $\times2$   & $1 \times 1$              & 30.82 & 0.8991      & 28.51 & 0.8446      & 28.39 & 0.8239       & 25.62 & 0.8199        & 28.12 & 0.9089        \\
                                      & Bilinear    & $\times2$   & $2 \times 2$             & 32.12 & 0.9106      & 29.15 & 0.8384      & 28.65 & 0.8090       & 25.95 & 0.8077        & 29.13 & 0.9115        \\
                                      & Bicubic     & $\times2$   & $4 \times 4$             & 33.63 & 0.9292      & 30.23 & 0.8681      & 29.53 & 0.8421       & 26.86 & 0.8394        & 30.78 & 0.9338        \\
                                      \midrule
     \multirow{3}{*}{LUT}          & SR-LUT \cite{DBLP:conf/cvpr/JoK21}    & $\times2$   & $3 \times 3$               & 35.46 & 0.9466      & 31.73 & 0.8958      & 30.64 & 0.8750       & 28.50 & 0.8777        & 33.87 & 0.9579        \\
                                     & MuLUT-SDY      & $\times2$   & $5 \times 5$             & 36.43 & 0.9530      & 32.35 & 0.9049      & 31.12 & 0.8849       & 29.10 & 0.8880        & 35.32 & 0.9656        \\
                                     & MuLUT-SDY-X2    & $\times2$   & $9 \times 9$               & \textbf{36.65} & \textbf{0.9541}      & \textbf{32.49} & \textbf{0.9065}      & \textbf{31.23} & \textbf{0.8865}       & \textbf{29.31} & \textbf{0.8910}        & \textbf{35.78} & \textbf{0.9674}       \\
                                     \midrule
     \multirow{4}{*}{Sparse   coding} & NE + LLE \cite{DBLP:conf/cvpr/ChangYX04}    & $\times2$   & -    & 35.79 & 0.9491      & 31.82 & 0.8996      & 30.77 & 0.8787       & 28.48 & 0.8803        & 33.95 & 0.9590        \\
                                      & Zeyde et al. \cite{DBLP:conf/cas/ZeydeEP10}   & $\times2$   & -    & 35.79 & 0.9494      & 31.87 & 0.8989      & 30.77 & 0.8771       & 28.47 & 0.8794        & 34.06 & 0.9599        \\
                                      & ANR \cite{DBLP:conf/iccv/TimofteDG13}   & $\times2$   & -          & 35.85 & 0.9500      & 31.86 & 0.9006      & 30.82 & 0.8800       & 28.49 & 0.8807        & 33.94 & 0.9597        \\
                                      & A+ \cite{DBLP:conf/accv/TimofteSG14}    & $\times2$   & -          & 36.57 & 0.9545      & 32.34 & 0.9056      & 31.21 & 0.8860       & 29.23 & 0.8938        & 35.32 & 0.9670        \\
                                      \midrule
     \multirow{4}{*}{DNN}             & FSRCNN \cite{DBLP:conf/eccv/DongLT16}    & $\times2$   & $17 \times 17$       & 37.05 & 0.9560      & 32.66 & 0.9090      & 31.53 & 0.8902       & 29.88 & 0.9020        & 36.67 & 0.9710 \\ 
                                      & CARN-M \cite{DBLP:conf/eccv/AhnKS18}   & $\times2$   & $45 \times 45$       & 37.42 & 0.9583      & 33.17 & 0.9136      & 31.88 & 0.8960       & 31.23 & 0.9192        & 37.60 & 0.9740        \\
                                      & RCAN \cite{DBLP:conf/eccv/ZhangLLWZF18}  & $\times2$   & Global         & 38.30 & 0.9617      & 34.14 & 0.9235      & 32.41 & 0.9025       & 33.17 & 0.9377        & 39.60 & 0.9791       \\
                                      & SwinIR \cite{DBLP:conf/cvpr/ZhangTKZ018}  & $\times2$   & Global & 38.14 & 0.9611 & 33.86 & 0.9206 & 32.31 & 0.9012 & 32.76 & 0.9340 & 39.12 & 0.9783\\
                                      \midrule \midrule
     \multirow{3}{*}{Interpolation}   & Nearest    & $\times3$   & $1 \times 1$             & 27.93 & 0.8123      & 26.00 & 0.7330      & 26.17 & 0.7065       & 23.34 & 0.6992        & 25.04 & 0.8157        \\
                                      & Bilinear   & $\times3$   & $2 \times 2$              & 29.54 & 0.8504      & 26.96 & 0.7526      & 26.77 & 0.7177       & 23.99 & 0.7135        & 26.15 & 0.8372        \\
                                      & Bicubic    & $\times3$   & $4 \times 4$              & 30.40 & 0.8678      & 27.55 & 0.7736      & 27.20 & 0.7379       & 24.45 & 0.7343        & 26.94 & 0.8554        \\
                                      \midrule
     \multirow{3}{*}{LUT}          & SR-LUT \cite{DBLP:conf/cvpr/JoK21}    & $\times3$   & $3 \times 3$               & 31.95 & 0.8969      & 28.73 & 0.8057      & 27.92 & 0.7690       & 25.53 & 0.7750        & 29.32 & 0.8970        \\
                                     & MuLUT-SDY   & $\times3$   & $5 \times 5$                & 32.59 & 0.9065      & 29.25 & 0.8194      & 28.23 & 0.7824       & 25.94 & 0.7899        & 30.34 & 0.9112        \\
                                     & MuLUT-SDY-X2  & $\times3$   & $9 \times 9$                 & \textbf{32.75} & \textbf{0.9089}      & \textbf{29.34} & \textbf{0.8215}      & \textbf{28.31} & \textbf{0.7841}       & \textbf{26.10} & \textbf{0.7945}        & \textbf{30.72} & \textbf{0.9161}       \\
                                     \midrule
     \multirow{4}{*}{Sparse   coding} & NE + LLE \cite{DBLP:conf/cvpr/ChangYX04}   & $\times3$   & -     & 31.87 & 0.8958      & 28.64 & 0.8085      & 27.92 & 0.7727       & 25.41 & 0.7755        & 28.70 & 0.8889        \\
                                      & Zeyde et al. \cite{DBLP:conf/cas/ZeydeEP10}  & $\times3$   & -     & 31.93 & 0.8969      & 28.70 & 0.8079      & 27.95 & 0.7715       & 25.45 & 0.7761        & 28.85 & 0.8920        \\
                                      & ANR \cite{DBLP:conf/iccv/TimofteDG13}  & $\times3$   & -           & 31.95 & 0.8970      & 28.69 & 0.8102      & 27.96 & 0.7745       & 25.45 & 0.7768        & 28.78 & 0.8900        \\
                                      & A+ \cite{DBLP:conf/accv/TimofteSG14}   & $\times3$   & -           & 32.63 & 0.9090      & 29.16 & 0.8190      & 28.28 & 0.7832       & 26.04 & 0.7974        & 29.90 & 0.9099        \\
                                      \midrule
     \multirow{4}{*}{DNN}             & FSRCNN \cite{DBLP:conf/eccv/DongLT16}   & $\times3$   & $17 \times 17$        & 33.18 & 0.9140      & 29.37 & 0.8240            & 28.53 & 0.7910       & 26.43 & 0.8080        & 31.10 & 0.9210    \\  
                                      & CARN-M \cite{DBLP:conf/eccv/AhnKS18}   & $\times3$   & $45 \times 45$       & 34.00 & 0.9235      & 29.99 & 0.8357      & 28.90 & 0.8001       & 27.55 & 0.8384        & 32.82 & 0.9385        \\
                                      & RCAN \cite{DBLP:conf/eccv/ZhangLLWZF18}  & $\times3$   & Global         & 34.78 & 0.9299      & 30.63 & 0.8477      & 29.33 & 0.8107       & 29.02 & 0.8695        & 34.58 & 0.9502     \\  
                                    & SwinIR \cite{DBLP:conf/cvpr/ZhangTKZ018}  & $\times3$   & Global         &34.62 &0.9289 &30.54 &0.8463 &29.20 &0.8082 &28.66 &0.8624 &33.98 &0.9478 \\
                                      \midrule \midrule
    \multirow{3}{*}{Interpolation} & Nearest                                    & $\times4$   & $1 \times 1$                     & 26.25 & 0.7372      & 24.65 & 0.6529      & 25.03 & 0.6293       & 22.17 & 0.6154        & 23.45 & 0.7414        \\
                                   & Bilinear                                    & $\times4$  & $2 \times 2$                      & 27.55 & 0.7884      & 25.42 & 0.6792      & 25.54 & 0.6460       & 22.69 & 0.6346        & 24.21 & 0.7666        \\
                                   & Bicubic                                      & $\times4$  & $4 \times 4$                      & 28.42 & 0.8101      & 26.00 & 0.7023      & 25.96 & 0.6672       & 23.14 & 0.6574        & 24.91 & 0.7871        \\ \midrule
    \multirow{4}{*}{LUT}        & SR-LUT \cite{DBLP:conf/cvpr/JoK21}        & $\times4$  & $3 \times 3$              & 29.82 & 0.8478      & 27.01 & 0.7355      & 26.53 & 0.6953       & 24.02 & 0.6990        & 26.80 & 0.8380        \\
                                    & MuLUT-SDY                          & $\times4$  & $5 \times 5$       & 30.40 & 0.8600      & 27.48 & 0.7507      & 26.79 & 0.7088       & 24.31 & 0.7137        & 27.52 & 0.8551        \\
                                    & MuLUT-SDY-X2                          & $\times4$ & $9 \times 9$       & \textbf{30.60} & \textbf{0.8653}      & \textbf{27.60} & \textbf{0.7541}      & \textbf{26.86} & \textbf{0.7110}      & \textbf{24.46} & \textbf{0.7194}      & \textbf{27.90} & \textbf{0.8633}   \\
                                    \midrule
    \multirow{4}{*}{Sparse coding} & NE + LLE \cite{DBLP:conf/cvpr/ChangYX04}                         & $\times4$   & -               & 29.62 & 0.8404      & 26.82 & 0.7346      & 26.49 & 0.6970       & 23.84 & 0.6942        & 26.10 & 0.8195        \\
                                   & Zeyde et al. \cite{DBLP:conf/cas/ZeydeEP10}                    & $\times4$  & -               & 26.69 & 0.8429      & 26.90 & 0.7354      & 26.53 & 0.6968       & 23.90 & 0.6962        & 26.24 & 0.8241        \\
                                   & ANR \cite{DBLP:conf/iccv/TimofteDG13}                            & $\times4$ & -               & 29.70 & 0.8422      & 26.86 & 0.7368      & 26.52 & 0.6992       & 23.89 & 0.6964        & 26.18 & 0.8214        \\
                                   & A+ \cite{DBLP:conf/accv/TimofteSG14}                             & $\times4$   & -              & 30.27 & 0.8602      & 27.30 & 0.7498      & 26.73 & 0.7088       & 24.33 & 0.7189        & 26.91 & 0.8480        \\ \midrule
    \multirow{4}{*}{DNN}           & FSRCNN \cite{DBLP:conf/eccv/DongLT16}                            & $\times4$  & $17 \times 17$                   & 30.72 & 0.8660      & 27.61 & 0.7550      & 26.98 & 0.7150       & 24.62 & 0.7280        & 27.90 & 0.8610       \\ 
                                   & CARN-M \cite{DBLP:conf/eccv/AhnKS18}                              & $\times4$ & $45 \times 45$                 & 31.82 & 0.8898      & 28.29 & 0.7747      & 27.42 & 0.7305       & 25.62 & 0.7694        & 29.85 & 0.8993        \\
                                   & RCAN \cite{DBLP:conf/eccv/ZhangLLWZF18}                            & $\times4$  & Global               & 32.61 & 0.8999      & 28.93 & 0.7894      & 27.80 & 0.7436       & 26.85 & 0.8089        & 31.45 & 0.9187        \\ 
                                   & SwinIR \cite{DBLP:conf/cvpr/ZhangTKZ018}                            & $\times4$  & Global                &32.44 &0.8976 &28.77 &0.7858 &27.69 &0.7406 &26.47 &0.7980 &30.92 &0.9151 \\
                                   \bottomrule     
    \end{tabular}%
    }
    \begin{tablenotes}
    \item[1] The PSNR and SSIM values are computed at the Y-channel in the YCbCr color space. 
    \end{tablenotes}

    \end{table*}

%% file: pics/visual_sr_main.tex
\newcommand{\name}{0}
\newcommand{\h}{0}
\newcommand{\w}{0.15}
\newcommand{\hrs}{0.324}
\newcommand{\vsp}{0.3 mm}
\newlength \g
%
\begin{figure*}[tp]
	\scriptsize
	\centering

\renewcommand{\name}{visual/X2/Set14/pepper/Y100_X190/pepper-}
\renewcommand{\h}{0.134}
\renewcommand{\w}{0.134}
\renewcommand{\hrs}{0.134}
\setlength{\g}{-4mm}
\begin{tabular}{cc}
	\hspace{-4mm}
	\begin{adjustbox}{valign=t}
		\begin{tabular}{c}
			\includegraphics[height=\hrs \textwidth, width=\hrs \textwidth]{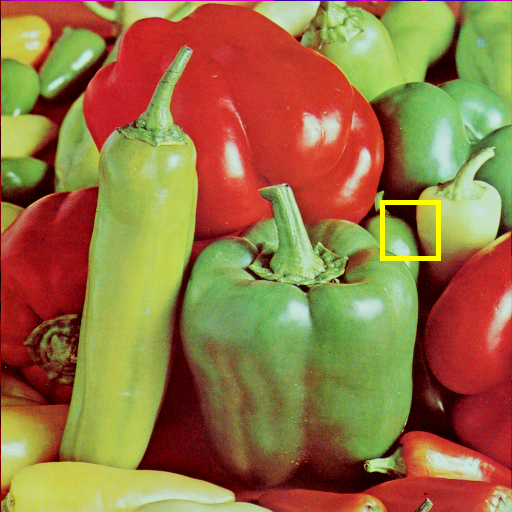}
			\\
			$\times 2$ super-resolution \\
			\textit{pepper} 
			from Set14
		\end{tabular}
	\end{adjustbox}
	\hspace{-1mm}
	\begin{adjustbox}{valign=t}
		\begin{tabular}{cccccc}
			\includegraphics[height=\h \textwidth, width=\w \textwidth]{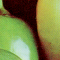} \hspace{\g} &
			\includegraphics[height=\h \textwidth, width=\w \textwidth]{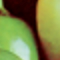} \hspace{\g} &
			\includegraphics[height=\h \textwidth, width=\w \textwidth]{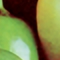} \hspace{\g} &
			\includegraphics[height=\h \textwidth, width=\w \textwidth]{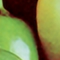}  \hspace{\g} &
			\includegraphics[height=\h \textwidth, width=\w \textwidth]{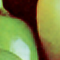}  \hspace{\g} &
			\includegraphics[height=\h \textwidth, width=\w \textwidth]{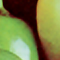}
			\\
			GT \hspace{\g} &
			Bicubic \hspace{\g} &
			A+ \hspace{\g} &
			FSRCNN \hspace{\g} &
			SR-LUT \hspace{\g} &
			Ours
			\\
			(PSNR/SSIM) \hspace{\g} &
			(35.03/0.9077) \hspace{\g} &
			(37.09/0.9227) \hspace{\g} &
			(37.06/0.9217) \hspace{\g} &
			(36.07/0.9170) \hspace{\g} &
			(36.56/0.9189)
			\\
		\end{tabular}
	\end{adjustbox}
\end{tabular}
\\
\vspace{\vsp}
\vspace{\vsp}
\vspace{\vsp}
\vspace{\vsp}
\vspace{\vsp}

\renewcommand{\name}{visual/X3/Urban100/img_092/Y65_X125/img_092-}
\renewcommand{\h}{0.134}
\renewcommand{\w}{0.134}
\renewcommand{\hrs}{0.134}
\setlength{\g}{-4mm}
\begin{tabular}{cc}
	\hspace{-4mm}
	\begin{adjustbox}{valign=t}
		\begin{tabular}{c}
			\includegraphics[height=\hrs \textwidth, width=\hrs \textwidth]{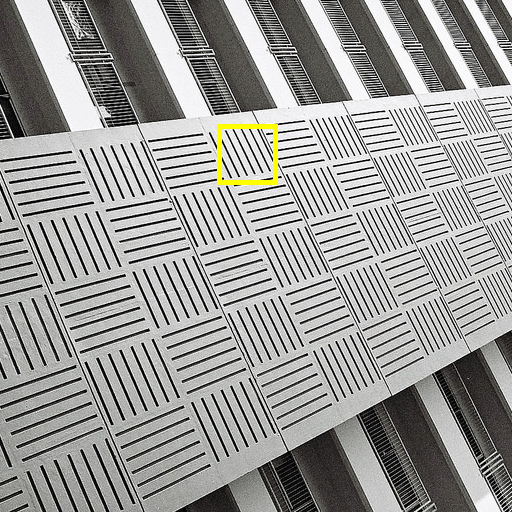}
			\\
			$\times 3$ super-resolution  \\
			\textit{img\_092}~(Urban100)
		\end{tabular}
	\end{adjustbox}
	\hspace{-1mm}
	\begin{adjustbox}{valign=t}
		\begin{tabular}{cccccc}
			\includegraphics[height=\h \textwidth, width=\w \textwidth]{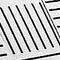} \hspace{\g} &
			\includegraphics[height=\h \textwidth, width=\w \textwidth]{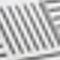} \hspace{\g} &
			\includegraphics[height=\h \textwidth, width=\w \textwidth]{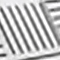} \hspace{\g} &
			\includegraphics[height=\h \textwidth, width=\w \textwidth]{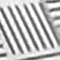} \hspace{\g} & 
			\includegraphics[height=\h \textwidth, width=\w \textwidth]{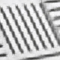} \hspace{\g} &
			\includegraphics[height=\h \textwidth, width=\w \textwidth]{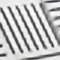}
			\\
			GT \hspace{\g} &
			Bicubic \hspace{\g} &
			A+ \hspace{\g} &
			FSRCNN \hspace{\g} &
			SR-LUT \hspace{\g} &
			Ours
			\\
			(PSNR/SSIM) \hspace{\g} &
			(17.29/0.5167) \hspace{\g} &
			(18.58/0.6331) \hspace{\g} &
			(18.52/0.6197) \hspace{\g} &
			(18.19/0.5993) \hspace{\g} &
			(18.46/0.6199)
            \\
		\end{tabular}
	\end{adjustbox}
\end{tabular}
\\
\vspace{\vsp}
\vspace{\vsp}
\vspace{\vsp}
\vspace{\vsp}
\vspace{\vsp}

			

\renewcommand{\name}{visual/X4/Manga109/TetsuSan/Y100_X100/TetsuSan-}
\renewcommand{\h}{0.134}
\renewcommand{\w}{0.134}
\renewcommand{\hrs}{0.134}
\setlength{\g}{-4mm}
\begin{tabular}{cc}
	\hspace{-4mm}
	\begin{adjustbox}{valign=t}
		\begin{tabular}{c}
			\includegraphics[height=\hrs \textwidth, width=\hrs \textwidth]{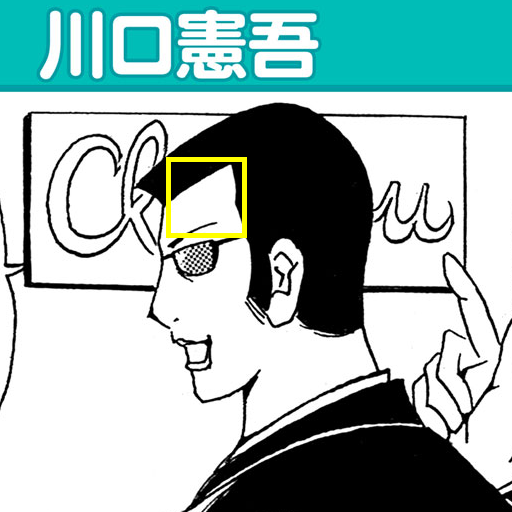}
			\\
			$\times 4$ super-resolution  \\
			\textit{TetsuSan}~(Manga109)

		\end{tabular}
	\end{adjustbox}
	\hspace{-1mm}
	\begin{adjustbox}{valign=t}
		\begin{tabular}{cccccc}
			\includegraphics[height=\h \textwidth, width=\w \textwidth]{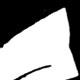} \hspace{\g} &
			\includegraphics[height=\h \textwidth, width=\w \textwidth]{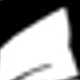} \hspace{\g} &
			\includegraphics[height=\h \textwidth, width=\w \textwidth]{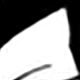} \hspace{\g} &
			\includegraphics[height=\h \textwidth, width=\w \textwidth]{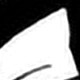} \hspace{\g} & 
			\includegraphics[height=\h \textwidth, width=\w \textwidth]{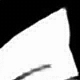} \hspace{\g} &
			\includegraphics[height=\h \textwidth, width=\w \textwidth]{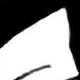}
			\\
			GT \hspace{\g} &
			Bicubic \hspace{\g} &
			A+ \hspace{\g} &
			FSRCNN \hspace{\g} &
			SR-LUT \hspace{\g} &
			Ours
			\\
			(PSNR/SSIM) \hspace{\g} &
			(20.72/0.8336) \hspace{\g} &
			(23.23/0.8997) \hspace{\g} &
			(24.36/0.9118) \hspace{\g} &
			(23.38/0.9045) \hspace{\g} &
			(24.75/0.9251)
            \\
		\end{tabular}
	\end{adjustbox}
\end{tabular}
\\
\vspace{\vsp}
\caption{ Visual comparison for $\times 2$, $\times 3$, and $\times 4$ image super-resolution on standard benchmark datasets. Best view in color and on screen.}
\label{fig:visual_sr_main}
\end{figure*}

%% file: pics/visual_dm_main.tex
%
\begin{figure*}[!t]
	\scriptsize
	\centering
%
%

\renewcommand{\name}{visual/dm/Kodak/21/Y202_X120/21-}
\renewcommand{\h}{0.134}
\renewcommand{\w}{0.134}
\renewcommand{\hrs}{0.300}
\setlength{\g}{-4mm}
\begin{tabular}{cc}
	\hspace{-4mm}
	\begin{adjustbox}{valign=t}
		\begin{tabular}{c}
			\includegraphics[height=\h \textwidth, width=\h \textwidth]{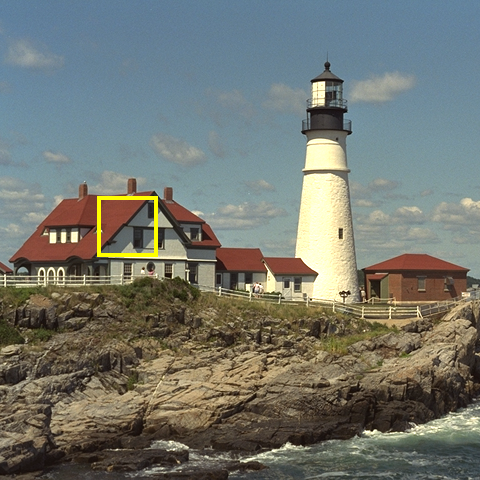}
			\\
			\textit{21} 
			(Kodak)
		\end{tabular}
	\end{adjustbox}
	\hspace{-1mm}
	\begin{adjustbox}{valign=t}
		\begin{tabular}{cccccc}
			\includegraphics[height=\h \textwidth, width=\w \textwidth]{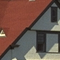}  \hspace{\g} &
			\includegraphics[height=\h \textwidth, width=\w \textwidth]{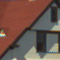} \hspace{\g} &
			\includegraphics[height=\h \textwidth, width=\w \textwidth]{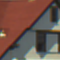} \hspace{\g} &
			\includegraphics[height=\h \textwidth, width=\w \textwidth]{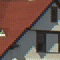} \hspace{\g} &
			\includegraphics[height=\h \textwidth, width=\w \textwidth]{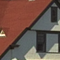} \hspace{\g} &
			\includegraphics[height=\h \textwidth, width=\w \textwidth]{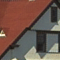}
			\\
			GT \hspace{\g} &
			Bilinear \hspace{\g} &
			Baseline-A \hspace{\g} &
			Baseline-B \hspace{\g} &
			DemosaicNet \hspace{\g} &
			Ours

			\\
		\end{tabular}
	\end{adjustbox} 
\end{tabular}

\vspace{\vsp} 

\caption{ Visual comparison for image demosaicing on the Kodak dataset. Best view in color and on screen.}
\label{fig:visual_dm_main}
\vspace{-4pt}
\end{figure*}

%% file: pics/fig_dm_tradeoff.tex
\begin{figure}[t]
    \begin{center}
        \includegraphics[width=0.9\columnwidth]{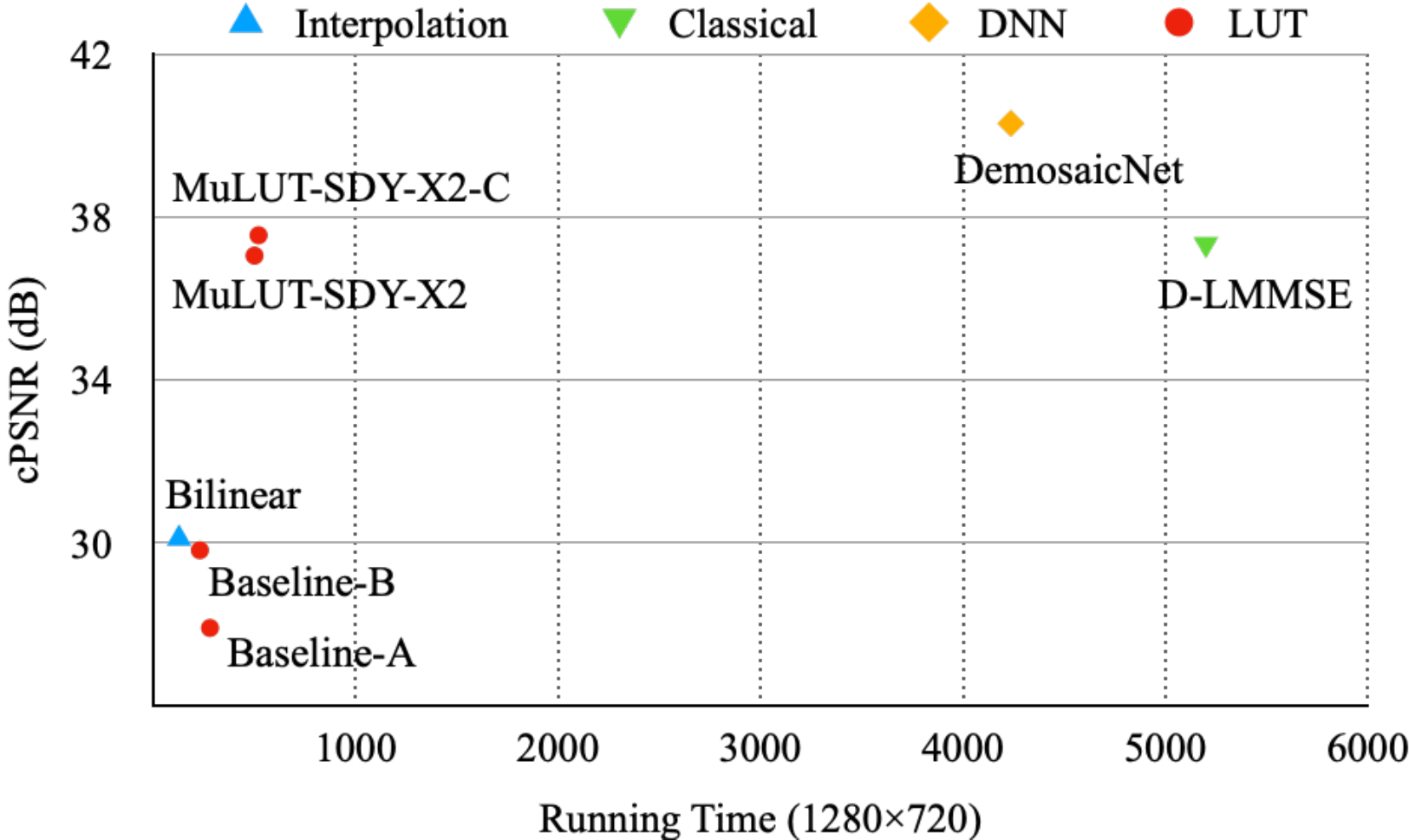}
    \end{center}
    \caption{The performance and running time tradeoff for the task of image demosaicing on the Kodak dataset. The cPSNR values are averaged over 3 color channels. The running time of classical solution is tested on a PC, and other methods are evaluated on a Xiaomi 11 mobile smartphone.} 
    \label{fig:dm_tradeoff}
\end{figure}

%% file: pics/visual_dn_main.tex
%
\begin{figure*}[tp]
	\scriptsize
	\centering
%

\renewcommand{\name}{visual/n15/test046/Y250_X200/test046-}
\renewcommand{\h}{0.134}
\renewcommand{\w}{0.134}
\renewcommand{\hrs}{0.134}
\setlength{\g}{-4mm}
\begin{tabular}{cc}
	\hspace{-4mm}
	\begin{adjustbox}{valign=t}
		\begin{tabular}{c}
			\includegraphics[height=\hrs \textwidth, width=\hrs \textwidth]{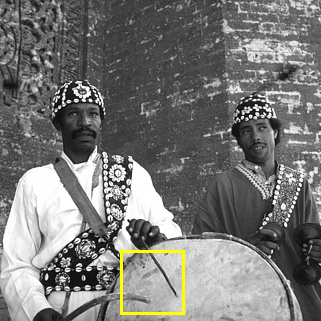}
			\\
			Noise Level = 15  \\
			\textit{test046}~(BSD68)
		\end{tabular}
	\end{adjustbox}
	\hspace{-1mm}
	\begin{adjustbox}{valign=t}
		\begin{tabular}{cccccc}
			\includegraphics[height=\h \textwidth, width=\w \textwidth]{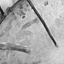} \hspace{\g} &
			\includegraphics[height=\h \textwidth, width=\w \textwidth]{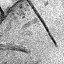} \hspace{\g} &
			\includegraphics[height=\h \textwidth, width=\w \textwidth]{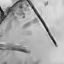} \hspace{\g} &
			\includegraphics[height=\h \textwidth, width=\w \textwidth]{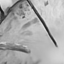} \hspace{\g} & 
			\includegraphics[height=\h \textwidth, width=\w \textwidth]{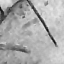} \hspace{\g} &
			\includegraphics[height=\h \textwidth, width=\w \textwidth]{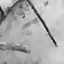}
			\\
			GT \hspace{\g} &
			Noisy \hspace{\g} &
			BM3D \hspace{\g} &
			DnCNN \hspace{\g} &
			SR-LUT \hspace{\g} &
			Ours
			\\
			(PSNR) \hspace{\g} &
			(24.75) \hspace{\g} &
			(28.56) \hspace{\g} &
			(29.13) \hspace{\g} &
			(28.19) \hspace{\g} &
			(29.13 )
            \\
		\end{tabular}
	\end{adjustbox}
\end{tabular}
\\
\vspace{\vsp}

\renewcommand{\name}{visual/n25/04/Y40_X80/04-}
\renewcommand{\h}{0.134}
\renewcommand{\w}{0.134}
\renewcommand{\hrs}{0.134}
\setlength{\g}{-4mm}
\begin{tabular}{cc}
	\hspace{-4mm}
	\begin{adjustbox}{valign=t}
		\begin{tabular}{c}
			\includegraphics[height=\hrs \textwidth, width=\hrs \textwidth]{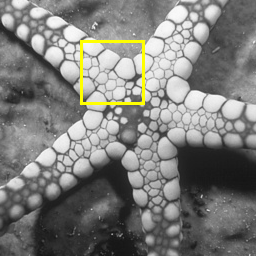}
			\\
			Noise Level = 25  \\
			\textit{04}~(Set12)
		\end{tabular}
	\end{adjustbox}
	\hspace{-1mm}
	\begin{adjustbox}{valign=t}
		\begin{tabular}{cccccc}
			\includegraphics[height=\h \textwidth, width=\w \textwidth]{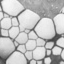} \hspace{\g} &
			\includegraphics[height=\h \textwidth, width=\w \textwidth]{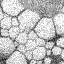} \hspace{\g} &
			\includegraphics[height=\h \textwidth, width=\w \textwidth]{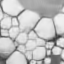} \hspace{\g} &
			\includegraphics[height=\h \textwidth, width=\w \textwidth]{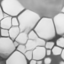} \hspace{\g} & 
			\includegraphics[height=\h \textwidth, width=\w \textwidth]{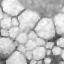} \hspace{\g} &
			\includegraphics[height=\h \textwidth, width=\w \textwidth]{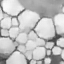}
			\\
			GT \hspace{\g} &
			Noisy \hspace{\g} &
			BM3D \hspace{\g} &
			DnCNN \hspace{\g} &
			SR-LUT \hspace{\g} &
			Ours
			\\
			(PSNR) \hspace{\g} &
			(20.44) \hspace{\g} &
			(28.64) \hspace{\g} &
			(29.32) \hspace{\g} &
			(26.94) \hspace{\g} &
			(28.21)
            \\
		\end{tabular}
	\end{adjustbox}
\end{tabular}
\\
\vspace{\vsp}
\renewcommand{\name}{visual/n50/07/Y40_X80/07-}
\renewcommand{\h}{0.134}
\renewcommand{\w}{0.134}
\renewcommand{\hrs}{0.134}
\setlength{\g}{-4mm}
\begin{tabular}{cc}
	\hspace{-4mm}
	\begin{adjustbox}{valign=t}
		\begin{tabular}{c}
			\includegraphics[height=\hrs \textwidth, width=\hrs \textwidth]{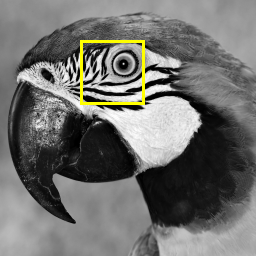}
			\\
			Noise Level = 50  \\
			\textit{07}~(Set12)
		\end{tabular}
	\end{adjustbox}
	\hspace{-1mm}
	\begin{adjustbox}{valign=t}
		\begin{tabular}{cccccc}
			\includegraphics[height=\h \textwidth, width=\w \textwidth]{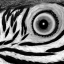} \hspace{\g} &
			\includegraphics[height=\h \textwidth, width=\w \textwidth]{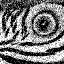} \hspace{\g} &
			\includegraphics[height=\h \textwidth, width=\w \textwidth]{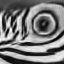} \hspace{\g} &
			\includegraphics[height=\h \textwidth, width=\w \textwidth]{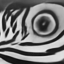} \hspace{\g} & 
			\includegraphics[height=\h \textwidth, width=\w \textwidth]{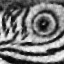} \hspace{\g} &
			\includegraphics[height=\h \textwidth, width=\w \textwidth]{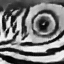}
			\\
			GT \hspace{\g} &
			Noisy \hspace{\g} &
			BM3D \hspace{\g} &
			DnCNN \hspace{\g} &
			SR-LUT \hspace{\g} &
			Ours
			\\
			(PSNR) \hspace{\g} &
			(15.08) \hspace{\g} &
			(25.01) \hspace{\g} &
			(25.04) \hspace{\g} &
			(22.43) \hspace{\g} &
			(25.24 )
            \\
		\end{tabular}
	\end{adjustbox}
\end{tabular}
\\
\vspace{\vsp}
\caption{ Visual comparison for image denoising with different noise levels on standard benchmark datasets. Best view on screen.}
\label{fig:visual_dn_main}
\end{figure*}

%% file: pics/visual_cdn_main.tex
%
\begin{figure*}[tp]
	\scriptsize
	\centering
%

\renewcommand{\name}{visual/cdn25/McMaster/18/Y30_X80/18-}
\renewcommand{\h}{0.134}
\renewcommand{\w}{0.134}
\renewcommand{\hrs}{0.134}
\setlength{\g}{-4mm}
\begin{tabular}{cc}
	\hspace{-4mm}
	\begin{adjustbox}{valign=t}
		\begin{tabular}{c}
			\includegraphics[height=\hrs \textwidth, width=\hrs \textwidth]{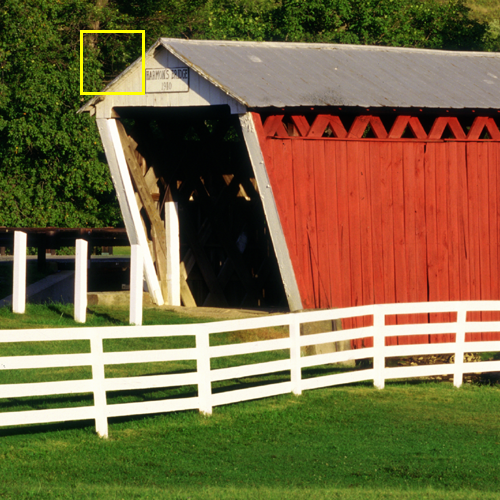}
			\\
			Noise Level = 25  \\
			\textit{18}~(McMaster)
		\end{tabular}
	\end{adjustbox}
	\hspace{-1mm}
	\begin{adjustbox}{valign=t}
		\begin{tabular}{cccccc}
			\includegraphics[height=\h \textwidth, width=\w \textwidth]{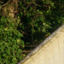} \hspace{\g} &
			\includegraphics[height=\h \textwidth, width=\w \textwidth]{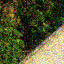} \hspace{\g} &
			\includegraphics[height=\h \textwidth, width=\w \textwidth]{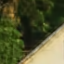} \hspace{\g} &
			\includegraphics[height=\h \textwidth, width=\w \textwidth]{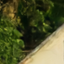} \hspace{\g} & 
			\includegraphics[height=\h \textwidth, width=\w \textwidth]{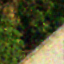} \hspace{\g} &
			\includegraphics[height=\h \textwidth, width=\w \textwidth]{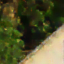}
			\\
			GT \hspace{\g} &
			Noisy \hspace{\g} &
			BM3D \hspace{\g} &
			DnCNN \hspace{\g} &
			SR-LUT \hspace{\g} &
			Ours
			\\
			(PSNR) \hspace{\g} &
			(20.95) \hspace{\g} &
			(29.69) \hspace{\g} &
			(31.19) \hspace{\g} &
			(27.50) \hspace{\g} &
			(29.25)
            \\
		\end{tabular}
	\end{adjustbox}
\end{tabular}
\\
\caption{Visual comparison for color image denoising on the McMaster dataset. Best view in color and on screen.}
\label{fig:visual_cdn_main}
\end{figure*}

%% file: tables/main_table_dn.tex

\begin{table}[t]
  \centering    
  \caption{The comparison in PSNR of different methods for grayscale image denoising on standard benchmark datasets under different noise levels, \emph{i.e.}, 15, 25, and 50. The performance of MuLUT exceeds SR-LUT significantly and approaches BM3D.}

    \resizebox{\columnwidth}{!}{%
    \begin{tabular}{llcccccc}
        \toprule
        \multirow{2}{*}{} & \multirow{2}{*}{Method}                    & \multicolumn{3}{c}{Set12}                               & \multicolumn{3}{c}{BSD68}                                                   \\ \cmidrule(lr){3-5} \cmidrule(lr){6-8}  
             &                                            & 15 & 25 & 50 & 15 & 25 & 50                        \\ \midrule
        \multirow{3}{*}{LUT}
           & SR-LUT \cite{DBLP:conf/cvpr/JoK21}       & 30.42 & 27.19 & 22.62 & 29.78 & 26.85 & 22.39                     \\
             & MuLUT-SDY-X2            & 31.50 & 28.94 & 25.46 & 30.63 & 28.18 & \textbf{24.97}                     \\
             & MuLUT-SDYEHO-X2                               & \textbf{31.77} & \textbf{29.18} & \textbf{25.47} & \textbf{30.89} & \textbf{28.34} & 24.96                     \\
             \midrule
        \multirow{3}{*}{Classical}            
           & BM3D \cite{DBLP:journals/tip/DabovFKE07}        & 32.37 & 29.97 & 26.72 & 31.07 & 28.57 & 25.62 \\
             & WNNM \cite{DBLP:conf/cvpr/GuZZF14}       & 32.70 & 30.26 & 27.05 & 31.37 & 28.83 & 25.87                     \\
             & TNRD \cite{DBLP:journals/pami/ChenP17}        & 32.50 & 30.06 & 26.81 & 31.42 & 28.92 & 25.97                     \\ \midrule
        \multirow{3}{*}{DNN}  & DnCNN \cite{DBLP:journals/tip/ZhangZCM017}       & 32.86 & 30.44 & 27.18 & 31.73 & 29.23 & 26.23                     \\
             & FFDNet \cite{DBLP:journals/tip/ZhangZZ18}       & 32.75 & 30.43 & 27.32 & 31.63 & 29.19 & 26.29                     \\
             & SwinIR \cite{DBLP:conf/eccv/WangYWGLDQL18} & 33.36 & 31.01 & 27.91 & 31.97 & 29.50 & 26.58                     \\ \bottomrule
    \end{tabular}%
    }
    \label{tab:main_dn}
    \end{table}

%% file: pics/fig_cdn_tradeoff.tex
\begin{figure}[t]
    \begin{center}
        \includegraphics[width=0.9\columnwidth]{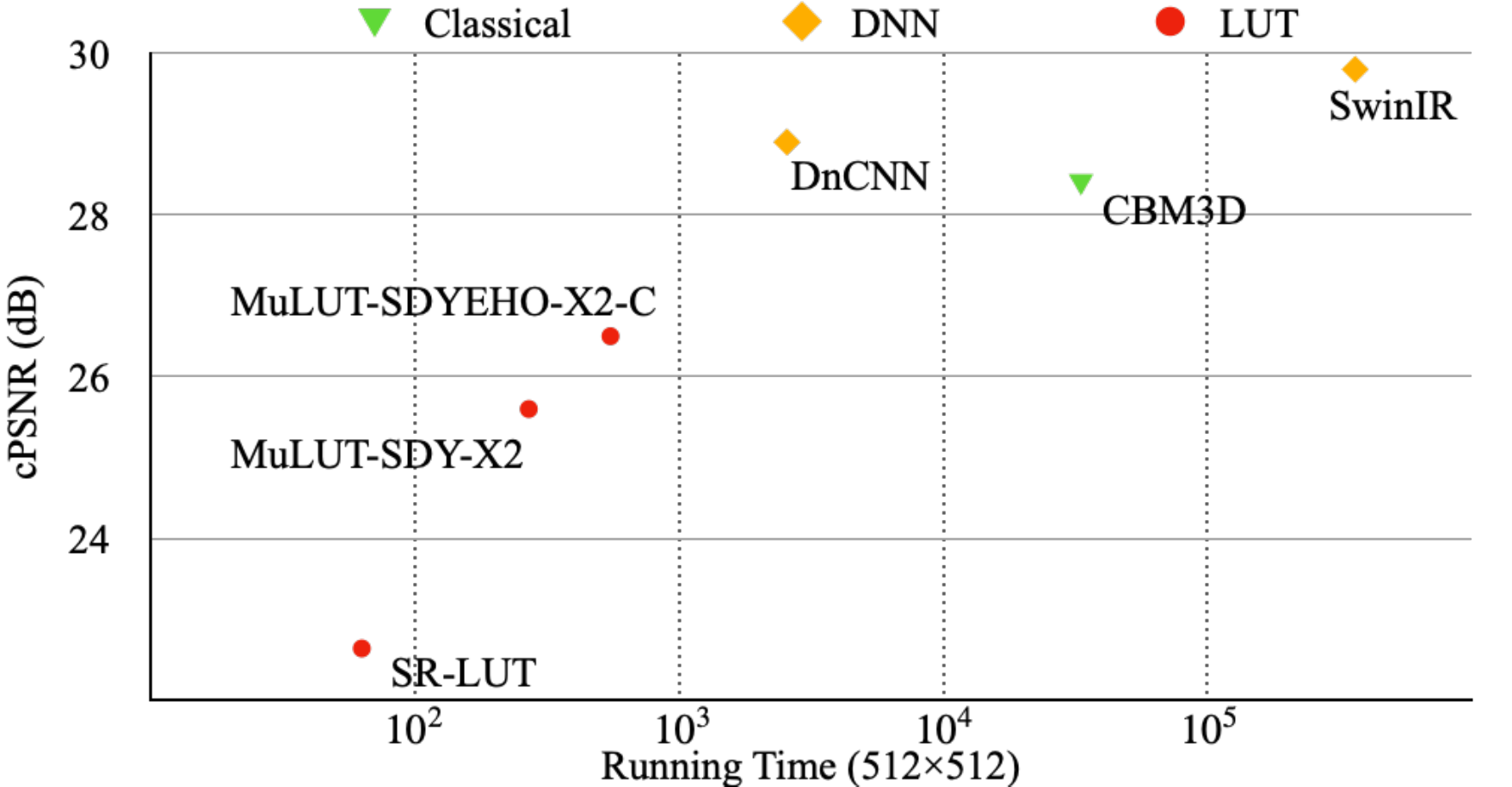}
    \end{center}
    \caption{The performance and running time tradeoff for the task of color image denoising on the Kodak24 dataset at a noise level of 50. The cPSNR values are averaged over 3 color channels. The running time of CBM3D is tested on a PC, and other methods are evaluated on a Xiaomi 11 mobile smartphone.} 
    \label{fig:cdn_tradeoff}
\end{figure}

%% file: pics/visual_db_main.tex
%
\begin{figure*}[tp]
	\scriptsize
	\centering
%

\renewcommand{\name}{visual/qf10/boats/Y290_X280/boats-}
\renewcommand{\h}{0.134}
\renewcommand{\w}{0.134}
\renewcommand{\hrs}{0.134}
\setlength{\g}{-4mm}
\begin{tabular}{cc}
	\hspace{-4mm}
	\begin{adjustbox}{valign=t}
		\begin{tabular}{c}
			\includegraphics[height=\hrs \textwidth, width=\hrs \textwidth]{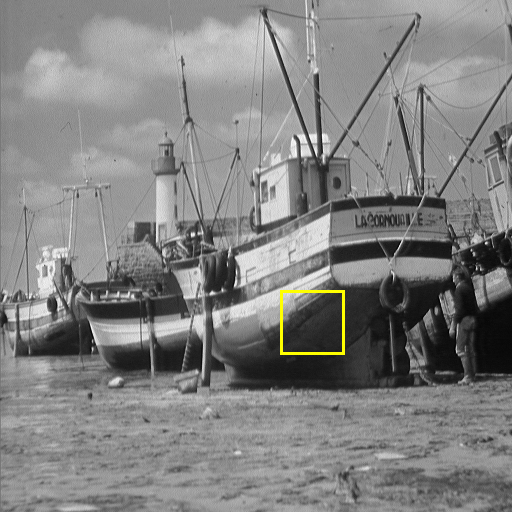}
			\\
			Quality Factor = 10  \\
			\textit{boats}~(classic5)
		\end{tabular}
	\end{adjustbox}
	\hspace{-1mm}
	\begin{adjustbox}{valign=t}
		\begin{tabular}{cccccc}
			\includegraphics[height=\h \textwidth, width=\w \textwidth]{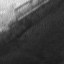} \hspace{\g} &
			\includegraphics[height=\h \textwidth, width=\w \textwidth]{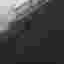} \hspace{\g} &
			\includegraphics[height=\h \textwidth, width=\w \textwidth]{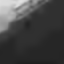} \hspace{\g} &
			\includegraphics[height=\h \textwidth, width=\w \textwidth]{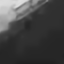} \hspace{\g} & 
			\includegraphics[height=\h \textwidth, width=\w \textwidth]{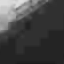} \hspace{\g} &
			\includegraphics[height=\h \textwidth, width=\w \textwidth]{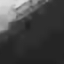}
			\\
			GT \hspace{\g} &
			JEPG \hspace{\g} &
			SA-DCT \hspace{\g} &
			ARCNN \hspace{\g} &
			SR-LUT \hspace{\g} &
			Ours
			\\
			(PSNR-B/SSIM) \hspace{\g} &
			(25.51/0.7580) \hspace{\g} &
			(28.47/0.7864) \hspace{\g} &
			(29.03/0.7950) \hspace{\g} &
			(27.92/0.7834) \hspace{\g} &
			(28.70/0.7890)
            \\
		\end{tabular}
	\end{adjustbox}
\end{tabular}
\\
\vspace{\vsp}

\renewcommand{\name}{visual/qf20/sailing2/Y330_X330/sailing2-}
\renewcommand{\h}{0.134}
\renewcommand{\w}{0.134}
\renewcommand{\hrs}{0.134}
\setlength{\g}{-4mm}
\begin{tabular}{cc}
	\hspace{-4mm}
	\begin{adjustbox}{valign=t}
		\begin{tabular}{c}
			\includegraphics[height=\hrs \textwidth, width=\hrs \textwidth]{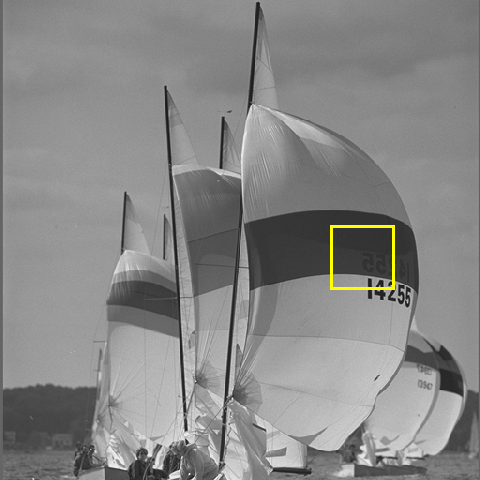}
			\\
			Quality Factor = 20  \\
			\textit{sailing2}~(LIVE1)
		\end{tabular}
	\end{adjustbox}
	\hspace{-1mm}
	\begin{adjustbox}{valign=t}
		\begin{tabular}{cccccc}
			\includegraphics[height=\h \textwidth, width=\w \textwidth]{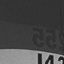} \hspace{\g} &
			\includegraphics[height=\h \textwidth, width=\w \textwidth]{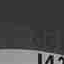} \hspace{\g} &
			\includegraphics[height=\h \textwidth, width=\w \textwidth]{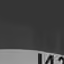} \hspace{\g} &
			\includegraphics[height=\h \textwidth, width=\w \textwidth]{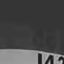} \hspace{\g} & 
			\includegraphics[height=\h \textwidth, width=\w \textwidth]{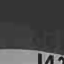} \hspace{\g} &
			\includegraphics[height=\h \textwidth, width=\w \textwidth]{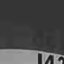}
			\\
			GT \hspace{\g} &
			JEPG \hspace{\g} &
			SA-DCT \hspace{\g} &
			ARCNN \hspace{\g} &
			SR-LUT \hspace{\g} &
			Ours
			\\
			(PSNR-B/SSIM) \hspace{\g} &
			(30.69/0.8914) \hspace{\g} &
			(33.41/0.9097) \hspace{\g} &
			(34.20/0.9129) \hspace{\g} &
			(32.77/0.9052) \hspace{\g} &
			(33.69/0.9102)
            \\
		\end{tabular}
	\end{adjustbox}
\end{tabular}
\\
\vspace{\vsp}

\caption{ Visual comparison for image deblocking with different quality factors on standard benchmark datasets. Best view on screen.}
\label{fig:visual_db_main}
\end{figure*}

%% file: tables/main_table_db.tex
\begin{table}[t]
    \caption{The comparison with other methods for image deblocking on standard benchmark datasets under different quality factors (QF). The performance is reported in the format of ``PSNR/SSIM/PSNR-B''. The performance of MuLUT exceeds SR-LUT significantly and approaches SA-DCT and ARCNN.}
    \centering
    \resizebox{\columnwidth}{!}{
        \begin{tabular}{llccc}        
        \toprule
                                & Method              & QF        & Classic5                & LIVE1                                                            \\ \midrule
        \multirow{2}{*}{Classical}       
& JPEG                       & 10 & 27.82/0.7595/25.21      & 27.77/0.7733/25.33                                               \\
& SA-DCT \cite{DBLP:journals/tip/FoiKE07}                    & 10 & 28.88/0.7894/28.15      & 28.65/0.7960/28.01                                               \\
\midrule
        \multirow{3}{*}{LUT}             
& SR-LUT \cite{DBLP:conf/cvpr/JoK21}                 & 10 & 28.59/0.7793/27.58      & 28.53/0.7948/27.69                                               \\
& MuLUT-SDY-X2              & 10 & \textbf{28.85}/0.7868/\textbf{28.29}      & 28.83/\textbf{0.8024}/28.39                                               \\
& MuLUT-SDYEHO-X2              & 10 & \textbf{28.85}/\textbf{0.7869}/28.26      & \textbf{28.86}/\textbf{0.8024}/\textbf{28.40}                                               \\
\midrule
        \multirow{2}{*}{DNN}             
& ARCNN  \cite{DBLP:conf/iccv/DongDLT15}                     & 10 & 29.03/0.7929/28.76      & 28.96/0.8076/28.77                                               \\
& SwinIR \cite{DBLP:conf/iccvw/LiangCSZGT21}                    & 10 & 30.27/0.8249/29.95      & 29.86/0.8287/29.50                                               \\ 
\midrule \midrule
        \multirow{2}{*}{Classical}       
& JPEG                       & 20 & 30.12/0.8344/27.50      & 30.08/0.8515/27.56                                               \\
& SA-DCT \cite{DBLP:journals/tip/FoiKE07}                     & 20 & 30.91/0.8488/29.75      & 30.81/0.8644/29.82                                               \\
\midrule
        \multirow{3}{*}{LUT}             
& SR-LUT \cite{DBLP:conf/cvpr/JoK21}                 & 20 & 30.83/0.8454/29.54      & 30.85/0.8658/29.74                                               \\
& MuLUT-SDY-X2             & 20 & 31.00/\textbf{0.8487}/30.30      & 31.11/0.8706/\textbf{30.52}                                               \\
& MuLUT-SDYEHO-X2              & 20 & \textbf{31.01}/0.8486/\textbf{30.31}      & \textbf{31.16}/\textbf{0.8707}/30.50                                               \\
\midrule
        \multirow{2}{*}{DNN}             
& ARCNN  \cite{DBLP:conf/iccv/DongDLT15}                     & 20 & 31.15/0.8517/30.59      & 31.29/0.8733/30.79                                               \\
& SwinIR \cite{DBLP:conf/iccvw/LiangCSZGT21}                     & 20 & 32.52/0.8748/31.99      & 32.25/0.8909/31.70                                               \\ 
\midrule \midrule
\multirow{2}{*}{Classical}       
& JPEG                       & 30 & 31.48/0.8666/28.94      & 31.41/0.8854/28.92                                               \\
& SA-DCT \cite{DBLP:journals/tip/FoiKE07}                     & 30 & 32.13/0.8754/30.82      & 32.08/0.8953/30.91                                               \\
\midrule
        \multirow{3}{*}{LUT}             
& SR-LUT \cite{DBLP:conf/cvpr/JoK21}                 & 30 & 32.13/0.8741/30.80      & 32.19/0.8963/30.99                                               \\
& MuLUT-SDY-X2             & 30 & \textbf{32.30}/\textbf{0.8768}/\textbf{31.57}      & 32.45/0.9004/31.79                                               \\
& MuLUT-SDYEHO-X2              & 30 & 32.29/0.8766/\textbf{31.57}      & \textbf{32.51}/\textbf{0.9009}/\textbf{31.84}                                               \\
\midrule
        \multirow{2}{*}{DNN}             
& ARCNN \cite{DBLP:conf/iccv/DongDLT15}                     & 30 & 32.51/0.8806/31.98      & 32.91/0.8861/32.38                                               \\
& SwinIR \cite{DBLP:conf/iccvw/LiangCSZGT21}                    & 30 & 33.73/0.8961/33.03      & 33.69/0.9174/33.01                                               \\ 
\midrule \midrule
\multirow{2}{*}{Classical}       
& JPEG                       & 40 & 32.43/0.8849/29.92      & 32.36/0.9044/29.95                                               \\
& SA-DCT \cite{DBLP:journals/tip/FoiKE07}                     & 40 & 32.99/0.8907/31.58      & 32.97/0.9128/31.77                                               \\
\midrule
        \multirow{3}{*}{LUT}             
& SR-LUT \cite{DBLP:conf/cvpr/JoK21}                  & 40 & 33.02/0.8904/31.71      & 33.15/0.9134/32.00                                               \\
& MuLUT-SDY-X2            & 40 & 33.16/0.8924/32.32      & 33.41/0.9168/32.66                                               \\
& MuLUT-SDYEHO-X2              & 40 & \textbf{33.18}/\textbf{0.8932}/\textbf{32.47}      & \textbf{33.49}/\textbf{0.9182}/\textbf{32.84}                                               \\
\midrule
        \multirow{2}{*}{DNN}             
& ARCNN  \cite{DBLP:conf/iccv/DongDLT15}                     & 40 & 33.32/0.8953/32.79      & 33.63/0.9198/33.14                                               \\
& SwinIR \cite{DBLP:conf/iccvw/LiangCSZGT21}                    & 40 & 34.52/0.9082/33.66      & 34.67/0.9317/33.88                                               \\ \bottomrule

    \end{tabular}%
    }
    \label{tab:main_db}
    \end{table}

%% file: tables/abl_table_net.tex
\begin{table}[t]
  \renewcommand\arraystretch{1.1}
  \caption{Ablation study on the network capacity for the task of image super-resolution.}
  \centering
  \resizebox{\columnwidth}{!}{%
  \begin{tabular}{lrrr|ccc} \toprule
    & \multicolumn{1}{c}{\begin{tabular}[c]{@{}c@{}} \# Param. \end{tabular}} & \multicolumn{1}{c}{\begin{tabular}[c]{@{}c@{}}LUT \\Size(MB)\end{tabular}} & \multicolumn{1}{c}{\begin{tabular}[c]{@{}c@{}}RF \\Size\end{tabular}} & \multicolumn{1}{c}{\begin{tabular}[c]{@{}c@{}}Set5\\ \end{tabular}} & \multicolumn{1}{c}{\begin{tabular}[c]{@{}c@{}}Set14\\ \end{tabular}}  & \multicolumn{1}{c}{\begin{tabular}[c]{@{}c@{}}Manga\\109 \end{tabular}}   \\ \midrule
  SR-LUT(nf.=64)  & 18.0K  & 1.274   &$3 \times 3$  & 29.82 & 27.01  & 26.80         \\ 
  SR-LUT(nf.=128)        & 71.38K       & 1.274 &$3 \times 3$ & 29.86 & 27.08  & 26.86         \\     
  SR-LUT(nf.=256)        & 268.6K       &  1.274 &$3 \times 3$ & 29.85 & 27.08 & 26.89         \\     
  \midrule
  MuLUT-SDY-X2(nf.=32)       & 71.4K       & 4.062 &$9 \times 9$ & 30.56 & 27.58  & 27.82        \\     
  MuLUT-SDY-X2(nf.=128)        & 1022.6K       & 4.062 &$9 \times 9$ & 30.54 & 27.59 & 27.83         \\     
  MuLUT-SDY-X2 w/o dc.       & 105.1K       & 4.062 &$9 \times 9$ & 30.54 & 27.56  & 27.82        \\       
  MuLUT-SDY-X2(nf.=64)   & 265.6K       & 4.062 &$9 \times 9$ & \textbf{30.60} & \textbf{27.60}  & \textbf{27.90}        \\ 
  \bottomrule     
\end{tabular}%
}
  \begin{tablenotes}
    \item[1] nf. denotes the number of filters of the convolutional layers inside SRNet of SR-LUT or a MuLUT block. For MuLUT-SDY-X2, nf. is set to 64 by default.
    \item[1] dc. denotes the dense connection inside each MuLUT block.
  \end{tablenotes}
  \label{tab:net_abl}
  \end{table}


%% file: tables/abl_table_patterns.tex

\begin{table}[t] \footnotesize
  \renewcommand\arraystretch{1.1}
  \centering
  \caption{Ablation study on the effectiveness of complementary indexing for the task of image super-resolution.}
  \resizebox{\columnwidth}{!}{%
  \begin{tabular}{lrrr|ccc} \toprule
    \qquad\qquad\qquad\qquad\qquad& \multicolumn{1}{c}{\begin{tabular}[c]{@{}c@{}}Energy \\Cost(pJ)\end{tabular}} & \multicolumn{1}{c}{\begin{tabular}[c]{@{}c@{}}LUT \\Size(MB)\end{tabular}} & \multicolumn{1}{c}{\begin{tabular}[c]{@{}c@{}}RF \\Size\end{tabular}} & \multicolumn{1}{c}{\begin{tabular}[c]{@{}c@{}}Set5\\ \end{tabular}} & \multicolumn{1}{c}{\begin{tabular}[c]{@{}c@{}}Set14\\ \end{tabular}}& \multicolumn{1}{c}{\begin{tabular}[c]{@{}c@{}}Manga\\109 \end{tabular}}   \\ \midrule
    MuLUT-S & 72.5M  & 1.274   &$3 \times 3$ & 29.82 & 27.01 & 26.80        \\
    MuLUT-SSS        & 222.3M       & 3.823 &$3 \times 3$ & 29.94 & 27.20 & 26.96        \\
    MuLUT-SD        & 149.2M       & 2.549 &$5 \times 5$ & 30.31 & 27.41  & 27.38        \\
    MuLUT-SDT      & 222.3M       & 3.823 &$5 \times 5$ & 30.35 & 27.45  & 27.50        \\ 
    MuLUT-SDY      & 222.3M       & 3.823 &$5 \times 5$ & 30.40 & 27.48  & 27.52        \\ 
    MuLUT-SDYEHO   & 441.6M        & 7.647 &$7 \times 7$ & 30.45 & 27.49  & 27.53 \\            
    MuLUT-SDY-X2    & 233.6M       & 4.062 &$9 \times 9$ & 30.60 & 27.60  & \textbf{27.90} \\           
    MuLUT-SDYEHO-X2   & 467.0M       & 8.124  &$13 \times 13$ & \textbf{30.62} & \textbf{27.61}  & 27.80  \\     
    \bottomrule     
  \end{tabular}%
  }
  \begin{tablenotes}
    \item[1] The energy cost is estimated for generating a $1280 \times 720$ HD image through $4 \times$ super-resolution.
  \end{tablenotes}
  \label{tab:p_abl}
  \end{table}


%% file: tables/abl_table_hierarchical.tex

\begin{table}[t] \footnotesize
  \renewcommand\arraystretch{1.1}
  \centering
  \caption{Ablation study on the effectiveness of hierarchical indexing for the task of image super-resolution.}
  \resizebox{\columnwidth}{!}{%
  \begin{tabular}{lrrr|ccc} \toprule
    & \multicolumn{1}{c}{\begin{tabular}[c]{@{}c@{}}Energy \\Cost(pJ)\end{tabular}} & \multicolumn{1}{c}{\begin{tabular}[c]{@{}c@{}}LUT \\Size(MB)\end{tabular}} & \multicolumn{1}{c}{\begin{tabular}[c]{@{}c@{}}RF \\Size\end{tabular}} & \multicolumn{1}{c}{\begin{tabular}[c]{@{}c@{}}Set5\\ \end{tabular}} & \multicolumn{1}{c}{\begin{tabular}[c]{@{}c@{}}Set14\\ \end{tabular}}  & \multicolumn{1}{c}{\begin{tabular}[c]{@{}c@{}}Manga\\109 \end{tabular}}   \\ \midrule
  MuLUT-S & 72.5M  & 1.274  &$3 \times 3$ & 29.82 & 27.01  & 26.80        \\
  MuLUT-S-X2 w/o ri. & 78.0M       & 1.354 &$5 \times 5$ & 30.11 & 27.26  & 27.02        \\ 
  MuLUT-S-X2   & 78.0M       & 1.354 &$5 \times 5$ & 30.23 & 27.39  & 27.39        \\                                    
  MuLUT-S-X3 & 83.4M       & 1.434 &$7 \times 7$ & 30.31 & 27.42  & 27.54        \\                                    
  MuLUT-S-X4   & 88.9M       & 1.513 &$9 \times 9$& \textbf{30.40} & \textbf{27.47}  & \textbf{27.66}        \\ \bottomrule
\end{tabular}%
}
  \begin{tablenotes}
    \item[1] ri. denotes the LUT re-indexing mechanism.
  \end{tablenotes}
  \label{tab:h_abl}
  \end{table}

%% file: tables/abl_table_hybrid_dm.tex
\begin{table}[t] \footnotesize
  \renewcommand\arraystretch{1.1}
  \caption{Ablation study on the effectiveness of channel indexing for the task of image demosaicing.}
  \centering
  \resizebox{0.9\columnwidth}{!}{%
  \begin{tabular}{lrrr|cc} \toprule
     ~~~~~~~~~~~~~~~~ & \multicolumn{1}{c}{\begin{tabular}[c]{@{}c@{}}Energy \\Cost(pJ)\end{tabular}} & \multicolumn{1}{c}{\begin{tabular}[c]{@{}c@{}}LUT \\Size(MB)\end{tabular}} & \multicolumn{1}{c}{\begin{tabular}[c]{@{}c@{}}RF \\Size\end{tabular}} &  \multicolumn{1}{c}{\begin{tabular}[c]{@{}c@{}}Kodak\\ \end{tabular}} & \multicolumn{1}{c}{\begin{tabular}[c]{@{}c@{}}McMaster\\ \end{tabular}} \\ \midrule
    Baseline-A  & 269.9M  & 0.319    &$3 \times 3$  & 27.90 & 29.27       \\ 
    Baseline-B  & 222.7M  &  0.956   &$3 \times 3$  & 29.81 & 28.72      \\ 
    MuLUT-SDY-X2       & 995.9M       & 1.195 &$9 \times 9$ & 37.11 & 35.09    \\     
    MuLUT-SDY-X2-C   & 1064.0M       & 1.677 &$9 \times 9$ & \textbf{37.54} & \textbf{35.11}         \\ \bottomrule     
  \end{tabular}%
  }
  \begin{tablenotes}
    \item[1] The energy cost is estimated for processing a $1280 \times 720$ HD mosaiced image.
    \item[2] ``-C'' denotes MuLUT with channel indexing.
  \end{tablenotes}
  \label{tab:c_dm_abl}
  \end{table}

  



%% file: tables/abl_table_hybrid_cdn.tex
\begin{table}[t] \footnotesize
  \renewcommand\arraystretch{1.1}
  \caption{Ablation study on the effectiveness of channel indexing for the task of color image denoising under noise level 50.}
  \centering
  \resizebox{\columnwidth}{!}{%
  \begin{tabular}{lrrr|ccc} \toprule
     ~~~~~~~~~~~~~~~~ & \multicolumn{1}{c}{\begin{tabular}[c]{@{}c@{}}Energy \\Cost(pJ)\end{tabular}} & \multicolumn{1}{c}{\begin{tabular}[c]{@{}c@{}}LUT \\Size(MB)\end{tabular}} & \multicolumn{1}{c}{\begin{tabular}[c]{@{}c@{}}RF \\Size\end{tabular}} & \multicolumn{1}{c}{\begin{tabular}[c]{@{}c@{}}CBSD68\\ \end{tabular}} & \multicolumn{1}{c}{\begin{tabular}[c]{@{}c@{}}Kodak\\ \end{tabular}} & \multicolumn{1}{c}{\begin{tabular}[c]{@{}c@{}}McMaster\\ \end{tabular}} \\ \midrule
    SR-LUT  & 69.4M  &  0.080  &$3 \times 3$  & 22.45 & 22.64 & 23.41       \\ 
    MuLUT-SDY-X2     & 391.0M          & 0.478   &$9 \times 9$ & 24.95 & 25.64 & 26.38        \\     
    MuLUT-SDYEHO-X2   &   782.0M    & 0.956 &$13 \times 13$ & 25.07 & 25.80 & 26.57         \\ 
    MuLUT-SDYEHO-X2-C   & 801.4M      & 2.882  &$13 \times 13$ & \textbf{25.90} & \textbf{26.49} & \textbf{26.77}         \\ 
    \bottomrule     
  \end{tabular}%
  }
  \begin{tablenotes}
    \item[1] The energy cost is estimated for denoising a $512 \times 512$ color image.
    \item[2] ``-C'' denotes MuLUT with channel indexing.
  \end{tablenotes}
  \label{tab:c_cdn_abl}
  \end{table}





%% file: tables/abl_table_ft.tex

%
\begin{table}[t] \footnotesize
  \renewcommand\arraystretch{1.1}
  \centering
  \caption{Ablation study on the effectiveness of the LUT-aware finetune strategy for the task of image super-resolution.}
  \resizebox{\columnwidth}{!}{%
  \begin{tabular}{lrrr|ccc} \toprule
    ~~~~~~~~~~~~~~~~ & \multicolumn{1}{c}{\begin{tabular}[c]{@{}c@{}}Energy \\Cost(pJ)\end{tabular}} & \multicolumn{1}{c}{\begin{tabular}[c]{@{}c@{}}LUT \\Size(MB)\end{tabular}} & \multicolumn{1}{c}{\begin{tabular}[c]{@{}c@{}}RF \\Size\end{tabular}} & \multicolumn{1}{c}{\begin{tabular}[c]{@{}c@{}}Set5\\ \end{tabular}} & \multicolumn{1}{c}{\begin{tabular}[c]{@{}c@{}}Set14\\ \end{tabular}} & \multicolumn{1}{c}{\begin{tabular}[c]{@{}c@{}}Manga\\109 \end{tabular}} \\ \midrule
   SR-LUT  net.  & -  & -   &$3 \times 3$  & 29.88 & 27.14 & 26.86        \\
   SR-LUT (4bit) w/o ft. & 72.5M  & 1.274   &$3 \times 3$ & 29.82 & 27.01 & 26.80 \\                             
   SR-LUT (4bit) w/ ft.   & 72.5M  & 1.274   &$3 \times 3$  & 29.94 & 27.18  & 26.94        \\
   SR-LUT (3bit) w/o ft. & 72.5M  & 0.100   &$3 \times 3$ & 29.58 & 26.99  & 26.76 \\                             
   SR-LUT (3bit) w/ ft.   & 72.5M  & 0.100   &$3 \times 3$  & 29.87 & 27.13  & 26.85        \\
   MuLUT-SDY-X2  net.       & -       & - &$9 \times 9$ & \textbf{30.61} & \textbf{27.61} & \textbf{27.93}        \\     
   MuLUT-SDY-X2  w/o ft.        & 233.6M       & 4.062 &$9 \times 9$ & 30.52 & 27.55  & 27.83        \\     
   MuLUT-SDY-X2   & 233.6M       & 4.062 &$9 \times 9$ & 30.60 & 27.60  & 27.90        \\ \bottomrule     
 \end{tabular}%
 }

  \begin{tablenotes}
    \item[1] net. denotes the performance of the corresponding neural network, 4bit denotes the sampling interval is $2^4$ and 3bit the $2^5$, and ft. denotes the LUT-aware finetuning strategy.
  \end{tablenotes}
  \label{tab:ft_abl}
  \end{table}

%% file: parts/conclusion.tex
\section{Conclusion and Future Work}

In this work, we propose a universal method to learn multiple look-up tables from data for image restoration tasks. Our method overcomes the limitation of the receptive field of a single LUT, empowering LUTs to be constructed like a neural network. Extensive experiments on image super-resolution, denoising, deblocking, and demosaicing demonstrate that MuLUT achieves significant improvement in restoration performance over the single-LUT solution while preserving its efficiency. Overall, MuLUT takes a step toward DNN of LUTs, showing its versatility in representative image restoration tasks and practicality for deployment on edge devices.

Nevertheless, compared with modern DNNs, MuLUT is relatively shallow and simple. As shown in our experiments, keeping enlarging RF contributes to better performance. Exploring more elaborated designs of constructing LUTs to obtain larger RF is worth trying. Besides, combining MuLUT with non-local operations like attention mechanisms is promising, too. Another limitation of MuLUT lies in the lack of ability for temporal modelling. Although with the channel indexing and pyramid structure, pixels along the temporal dimension can be involved, it requires a lot of LUTs to obtain a large temporal window. Our future work would include extending MuLUT to video restoration tasks like video super-resolution and denoising.

